\documentclass[5p, times]{elsarticle}

\journal{Signal Processing: Image Communication}

\usepackage{graphicx}
\usepackage{amssymb}
\usepackage{amsmath}
\usepackage{mathtools}
\usepackage{latexsym}
\usepackage{url}
\usepackage{xcolor}
\usepackage{subfigure}
\usepackage{subfloat}
\usepackage{xcolor, colortbl}
\usepackage{multirow}
\usepackage{diagbox}
\usepackage{algorithm,algcompatible,amsmath}
\usepackage{etoolbox}
\algnewcommand\INPUT{\item[\textbf{Input:}]}
\algnewcommand\OUTPUT{\item[\textbf{Output:}]}
\DeclareMathOperator*{\argmax}{argmax}

\patchcmd{\emailauthor}{(#2)}{}{}{}
\patchcmd{\urlauthor}{(#2)}{}{}{}

\begin{document}

\begin{frontmatter}

\title{Learning deep features for source color laser printer identification \\ based on cascaded learning}

\author{Do-Guk Kim$^a$}
\author{Jong-Uk Hou$^b$}
\author{Heung-Kyu Lee$^{b,}$\corref{mycorrespondingauthor}}

\address{$^a$Graduate School of Information Security, KAIST, $^b$School of Computing, KAIST}
\address{Korea Advanced Institute of Science and Technology, Guseong-dong, Yuseong-gu, Daejeon 305-701, Republic of Korea}

\cortext[mycorrespondingauthor]{Corresponding author}
\ead{heunglee@kaist.ac.kr}

\begin{abstract}
Color laser printers have fast printing speed and high resolution, and forgeries using color laser printers can cause significant harm to society. A source printer identification technique can be employed as a countermeasure to those forgeries. This paper presents a color laser printer identification method based on cascaded learning of deep neural networks. The refiner network is trained by adversarial training to refine the synthetic dataset for halftone color decomposition. The halftone color decomposing ConvNet is trained with the refined dataset, and the trained knowledge is transferred to the printer identifying ConvNet to enhance the accuracy. The robustness about rotation and scaling is considered in training process, which is not considered in existing methods. Experiments are performed on eight color laser printers, and the performance is compared with several existing methods. The experimental results clearly show that the proposed method outperforms existing source color laser printer identification methods.
\end{abstract}

\begin{keyword}
Generative adversarial network \sep Convolutional neural network \sep Color laser printer \sep Source printer identification \sep Mobile camera
\end{keyword}

\end{frontmatter}


\section{Introduction}

The development of color laser printing makes printing much easier than before. However, powerful printing devices are often abused to make forged documents. Because of the high resolution of modern color laser printers, it is hard for ordinary people to distinguish forged documents from genuine documents. Moreover, a color laser printer has a fast printing speed so that those forgeries can be produced in large quantities. Banknote forgeries and document forgeries on a large scale can cause harm to the society.

To prevent those forgeries, researchers have introduced various source laser printer identification methods. Source laser printer identification is a multimedia forensic technique that can be employed as a countermeasure to forgeries made by laser printers. Counterfeiters use a laser printer different from the source printer of the genuine documents. Even if the model of their printer is the same as that of the source printer of the genuine documents, every printing device is uniquely different from all other printers. Source laser printer identification techniques identify the exact source printing device of the target printed material. It can be utilized to help forgery investigation, and it can also be used as a part of a forgery detection system.

\begin{figure}[t]
    \centerline{\includegraphics[width=9cm]{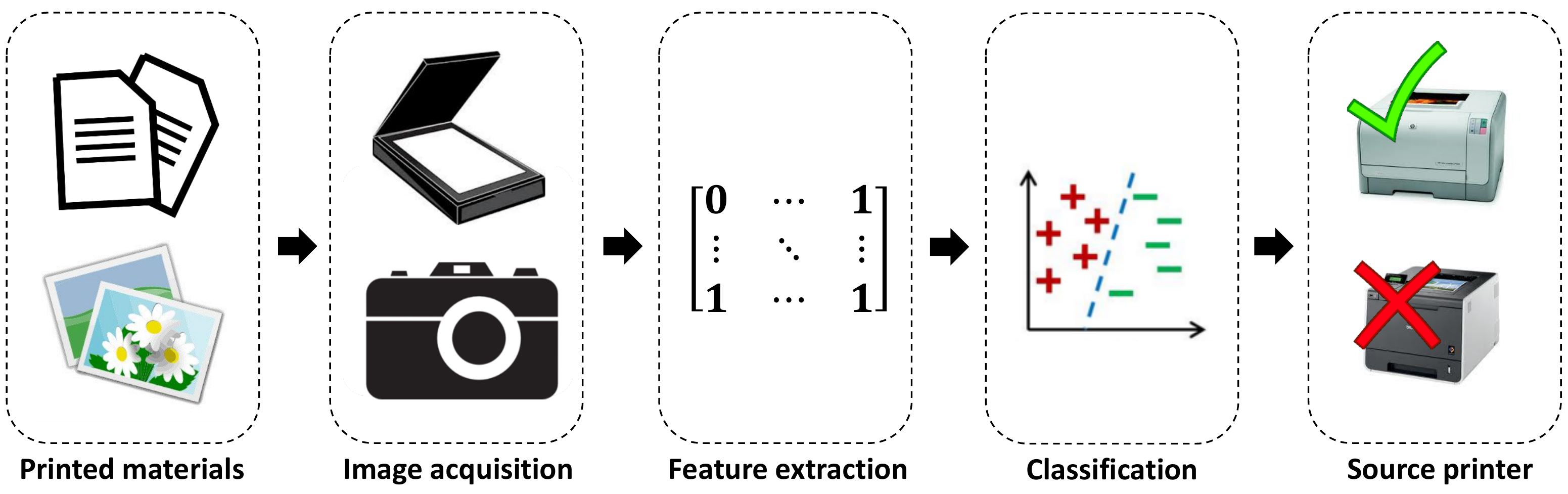}}
    \caption{ Source laser printer identification process
    } \label{PrinterIdentify}
\end{figure}

The typical process of source laser printer identification techniques is shown in Figure \ref{PrinterIdentify}. First, features are extracted from a scanned or photographed image of the target printed material. In this process, various image processing techniques are used to extract features from the image. Then, the features are used to classify the source printer of the target printed material. In the classification process, machine learning techniques are used, and reference features extracted from the known printed materials are used to train the classifier.

Majority of existing source laser printer identification techniques used scanned images as inputs. Scanning is a suitable method to acquire document image in a forensic investigation. However, these techniques have a limit to being distributed to the public since most scanners are not portable. While most scanners are not portable, mobile cameras are widespread due to the high distribution rate of smartphones. Therefore, to prevent document forgery effectively by using source printer identification techniques, it is essential to use photographed images as inputs.

There were several existing methods that used photographed images as inputs. However, they have a low applicability in two aspects. First, their identification accuracy is low to be applied to a real forensic application. Their low identification accuracy is mainly caused by the difficulty of halftone color channel decomposition. CMYK toners were used in the printing process while the digital image of the printed image is photographed with RGB color channels. Existing methods use CMY or CMYK color domain converted by using pre-defined color profile, but CMYK toner patterns are not decomposed clearly in the converted color domain. Thus, clear decomposition of CMYK toner patterns is needed to improve the identification accuracy. Second, the robustness against scaling and rotation is not considered in existing methods. Unlike scanning, it is difficult to keep the photographing distance and angle identically in photographing environment. Therefore, achieving robustness against scaling and rotation is necessary for printer identification with photographed images.

In this paper, we present a method based on cascaded learning of Generative Adversarial Networks (GANs) and Convolutional Neural Networks (CNNs) to identify the source color laser printer of photographed color documents. The proposed method is divided into two components; improved halftone color channel decomposition based on a GAN-alike framework, and printer identification based on a CNN. GAN framework was used to generate training data for halftone color channel decomposition. Since there is no label for decomposed toner channels of the photographed color images, we utilized the Simulated+Unsupervised (S+U) learning introduced in \cite{SimGan}. After that, a CNN was trained to decompose CMYK halftone color channels of input RGB image. Halftone color channel decomposition was carried out by the trained CNN. Our color decomposition method exceeds by far the performance of the existing color decomposition method which based on pre-defined color profiles.

Weights of the Halftone Color Decomposing-CNN (HCD-CNN) were used to initialize the Printer Identifying-CNN (PI-CNN). Knowledge about decomposing color channels of photographed color documents was transferred to the PI-CNN. Then, the PI-CNN was trained for two phases. The first phase was training for original input images, and the second phase was training for robustness about scaling and rotation. As a result of this two-step training, the PI-CNN not only showed a high identification accuracy but also achieved a robustness against various scaling and rotation values that were not trained. We performed experiments to verify the performance of the proposed method comparing it to several previous methods: Kim's methods (\cite{Kim1} and \cite{Kim2}), Tsai's method \cite{Tsai1}, and Ryu's method \cite{Ryu}. Thus, a total of five methods were used to compare the performance. For each method, the same printed materials were used in training and testing. The experimental results showed that the proposed method had overcome the limitations of the existing methods.

Major contributions of this paper are:

\begin{itemize}
\item We propose an improved halftone color channel decomposition method based on a GAN-alike framework using S+U learning.
\item We propose a color laser printer identification method based on a CNN which shows the state-of-the-art performance.
\item We achieve a robustness against rotation and scaling which is not considered in existing methods.
\end{itemize}

The rest of the paper is organized as follows. Section 2 describes some background on source color laser printer identification for photographed color documents. The presented method is described in Section 3. In Section 4, experimental results are reported. Section 5 gives our conclusions and presents the further research issues.

\section{Background}
\label{sec:2}
\subsection{Related work}
\subsubsection{Source laser printer identification methods}

Source laser printer identification techniques can be classified into two categories: methods for text documents and methods for color documents. After the Mikkilineni et al. \cite{Mik1} firstly suggested a source laser printer identification method using the banding frequency of the source printer,  researchers have proposed various methods that identify the source laser printer of the input text \cite{Mik2}-\cite{Ferreira2}. Most of them \cite{Mik2}\cite{Deng}\cite{Zhou}\cite{Ferreira1} are based on analyzing texture of printed letters. The texture is analyzed by using Gray-Level Co-occurrence Matrix (GLCM) related features. Bulan et al. \cite{Bulan} introduced a method based on the geometric distortion feature of the halftone printed dots. Geometric distortion features were extracted by estimating an ideal halftone dot position and subtracting it from the real halftone pattern. Recently, Ferreira et al. \cite{Ferreira2} proposed a data-driven source laser printer identification method that uses CNN as the classifier. It showed high accuracy rate on noisy text image data and outperforms existing other methods for text documents.

\begin{figure*}[t!]
    \centerline{\includegraphics[width=16cm]{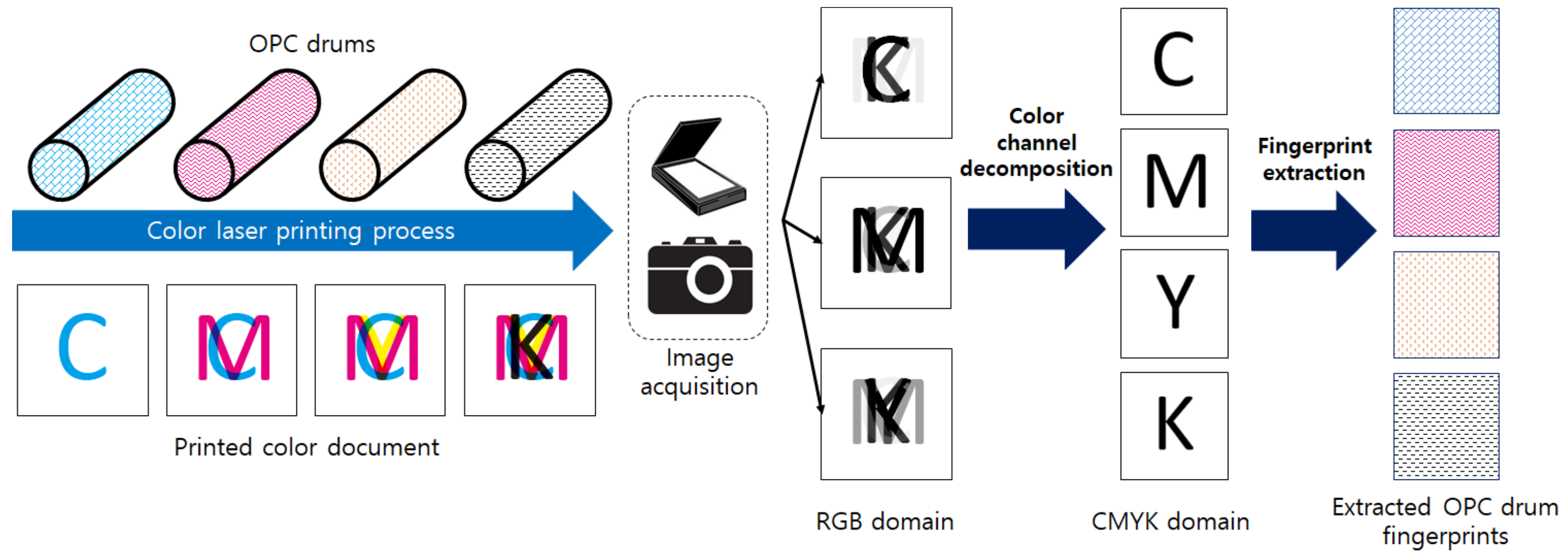}}
    \caption{Color laser printing process and halftone color channel decomposition
    } \label{PrintProcess}
\end{figure*}

Source color laser printer identification is a technique that extracts features from color laser printed images unlike methods for text documents. Choi et al. \cite{Choi1} proposed a source color laser printer identification method using noise features extracted from an HH band of the wavelet domain. Thirty-nine statistical features were extracted, and the support vector machine (SVM) was used as a classifier. They expanded their work as a method using GLCM of the color noise \cite{Choi2}. Statistical features were extracted from the GLCM of the color noise, and it was used to train and test the SVM.

Ryu et al. \cite{Ryu} suggested a method using a halftone printing angle histogram as feature vectors. A halftone printing angle histogram was extracted using the Hough transform in the CMY domain. Then, the source printer was identified by correlation-based detection. While Choi's method \cite{Choi1} used only the HH band of the wavelet domain, Tsai et al. \cite{Tsai1} introduced an identification method that utilized noise features from all bands of the wavelet domain. After that, they expanded their work as a hybrid identification method using both noise features from color images and GLCM features from monochrome characters \cite{Tsai2}. They adopted feature selection algorithms to find the best feature set. SVM-based classification was then used to identify the source printer.

While the printer identification methods mentioned above use scanned images, Kim and Lee \cite{Kim1} presented a method that using photographed images from a mobile device as input. Kim's method can identify a source color laser printer with photographed images; however, it required an additional close-up lens to acquire useful input images. Therefore, they suggested a method using a halftone texture fingerprint extracted from photographed images \cite{Kim2}. The method proposed in \cite{Kim2} used photographed images that do not require an additional close-up lens as input images. 

\subsubsection{Simulated+Unsupervised learning}

The GAN framework is composed of two competing networks. A generator generates a synthetic image similar with a real image, whereas a discriminator tries to classify whether the input is the real image or not. Goodfellow et al. \cite{Gan} firstly introduced the GAN framework, and many improvements and applications (such as EBGAN\cite{ebgan}, BEGAN\cite{began}, DCGAN\cite{dcgan}, SRGAN\cite{srgan}) have been presented.

Shrivastava et al. \cite{SimGan} suggested S+U learning based on the SimGAN framework. S+U learning means learning of a model to improve the realism of a simulated synthetic image using unlabeled real data while preserving the annotation information of the synthetic image. SimGAN is composed of refiner network and discriminator network. The refiner is similar to a generator of the traditional GAN framework, and simulated synthetic images are refined using the trained refiner network. In the training process, the refiner network and the discriminator network compete to reduce their losses. After the losses are stabilized, the trained refiner can refine synthetic image to be similar to a real image while preserving the annotation information.

\subsection{Halftone color channel decomposition}

Color laser printing process and importance of halftone color channel decomposition is described in Figure \ref{PrintProcess}. CMYK toners are printed by rolling of Optical PhotoConductor (OPC) drum in the toner cartridge, and OPC drum fingerprints are imported in printed toner pattern. OPC drum fingerprint means unique halftone pattern printed by the OPC drum. Each OPC drum has unique halftone pattern because even the OPC drums of same cartridge model have different geometric distortion or noise. Therefore, unique halftone pattern of the OPC drum or extracted features from the pattern can be used as a fingerprint of the OPC drum.

Since a digital image is photographed or scanned in RGB color domain,  CMYK toners are mixed and presented in all RGB channel as shown in Figure \ref{PrintProcess}. If the halftone color channel decomposing works perfect, we can extract OPC drum fingerprints from decomposed CMYK channel. However, existing color domain transition method couldn't clearly decompose each CMYK toner pattern to different channels. This is mainly caused by image processing arose in digital camera or scanner.

One possible way to decompose each CMYK toner pattern is a machine-learning based method. If we have photographed color halftone image that CMYK toner pattern for printing is known, we can use supervised learning. However, CMYK toner pattern for printing is hard to acquire because it is processed internally in printer driver or firmware. On the other way, generating synthetic color halftone image can be done easily in photo editing software such as Photoshop. Then, since we have synthetic images with annotation information and unlabeled real images, we can adopt S+U learning to create datasets for supervised learning. Therefore, we adopted S+U learning to create datasets for halftone color channel decomposition.

\subsection{Differences between photographing environment and scanning environment}

Most existing source color laser printer identification techniques use scanned images as input. These methods cannot be used to identify the source printer of input images photographed with mobile devices because of differences between the scanning environment and the photographing environment. While the intensity of the illumination is uniform in scanned images, it is not in photographed images. Moreover, blurring caused by defocusing can occur in photographed images, but this does not happen with scanned images.

\begin{figure}[t!]
    \centerline
    {
    	\subfigure[High resolution scan] {\includegraphics[width=4.2cm]{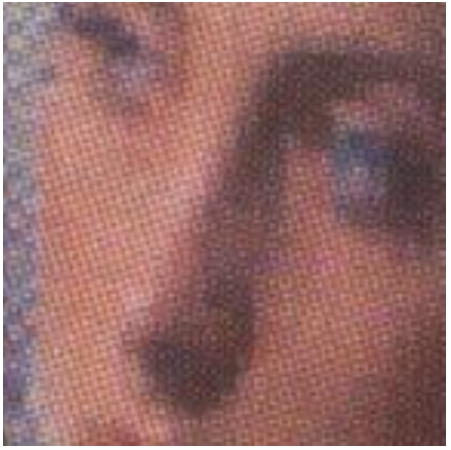}	\label{HR}}
    	    	\hspace{2pt}
	\subfigure[Low resolution scan]	{\includegraphics[width=4.2cm]{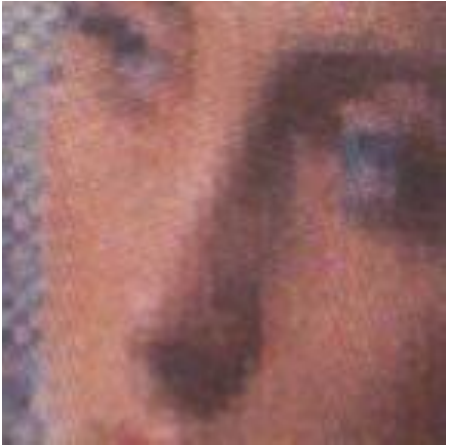}	\label{LR}}
    }
	\centerline{	
    	\subfigure[Close-up photography]	{\includegraphics[width=4.2cm]{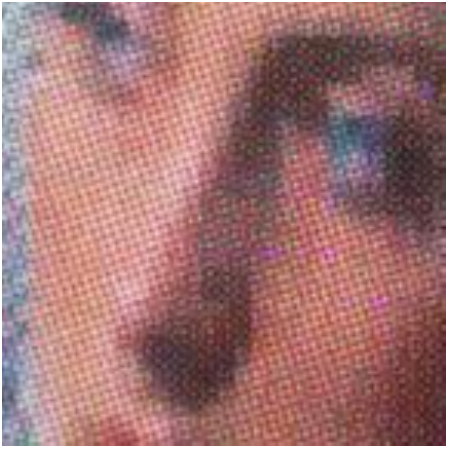}	\label{Closeup}}
    	    	\hspace{2pt}
	\subfigure[Normal photography]	{\includegraphics[width=4.2cm]{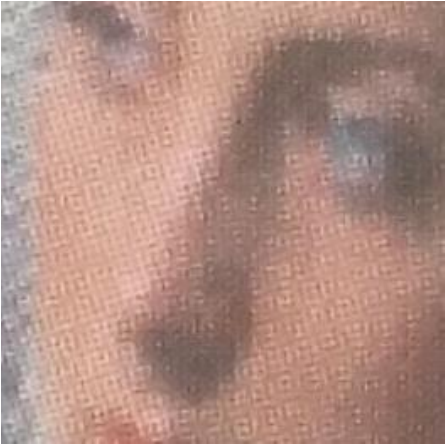}	\label{Normal}}
	}
    \caption{Various image acquisition results}
    \label{ImageAcquisition}	
\end{figure}

Comparison results of various image acquisition methods are shown in Figure \ref{ImageAcquisition}. Figure \ref{HR} is an image scanned at 1200 dpi, Figure \ref{LR} is an image scanned at 400 dpi, Figure \ref{Closeup} is an image photographed with a mobile device equipped with an additional close-up lens, and Figure \ref{Normal} is an image photographed with a normal mobile device. All images were acquired from the same printed image. Halftone dots appear clearly in high resolution images (Figure \ref{HR} and Figure \ref{Closeup}) while they do not in low resolution images (Figure \ref{LR} and Figure \ref{Normal}). The illumination is uniform in the scanned images (Figure \ref{HR} and Figure \ref{LR}) while it is not in the photographed images (Figure \ref{Closeup} and Figure \ref{Normal}).

Source color laser printer identification techniques that use scanned images as input images could not identify the source printer of photographed input images, for either close-up photographed images or normally photographed images. To overcome this limitation of the color laser printer identification techniques, Kim and Lee \cite{Kim1} presented a method that uses halftone texture features of photographed images. They extracted texture features from close-up photographed halftone images and used the extracted features to train and test the SVM. Adaptive thresholding was adopted to extract halftone patterns under non-uniform illumination of the photographed images.

In \cite{Kim1}, they used three sorts of halftone texture features: printing angle, printing resolution, and detail texture. However, it is difficult to extract the detail texture feature from normally photographed images. The method presented in \cite{Kim1} was not able to analyze normally photographed halftone images taken from mobile devices. Therefore, they extracted the halftone texture fingerprint from normally photographed images \cite{Kim2}, and they used it in source color laser printer identification. Halftone texture fingerprints were extracted in the discrete curvelet transform domain. Noise components were removed in the extraction process, and the extracted halftone texture fingerprint included printing angle features and printing resolution features.

Although the halftone texture fingerprint can be used in color laser printer identification, it has an inherent limitation in detection accuracy. Without detail texture features, it is hard to distinguish color printers made from the same manufacturer since the printing angle is similar. Therefore, we adopted a CNN to identify the source color laser printer of normally photographed images. All three sorts of halftone texture features are considered in feature learning process in a CNN. As a result, the CNN was able to identify source color laser printers that had similar halftone patterns by analyzing these features.

\begin{figure}[t!]
    \centerline{\includegraphics[width=9cm]{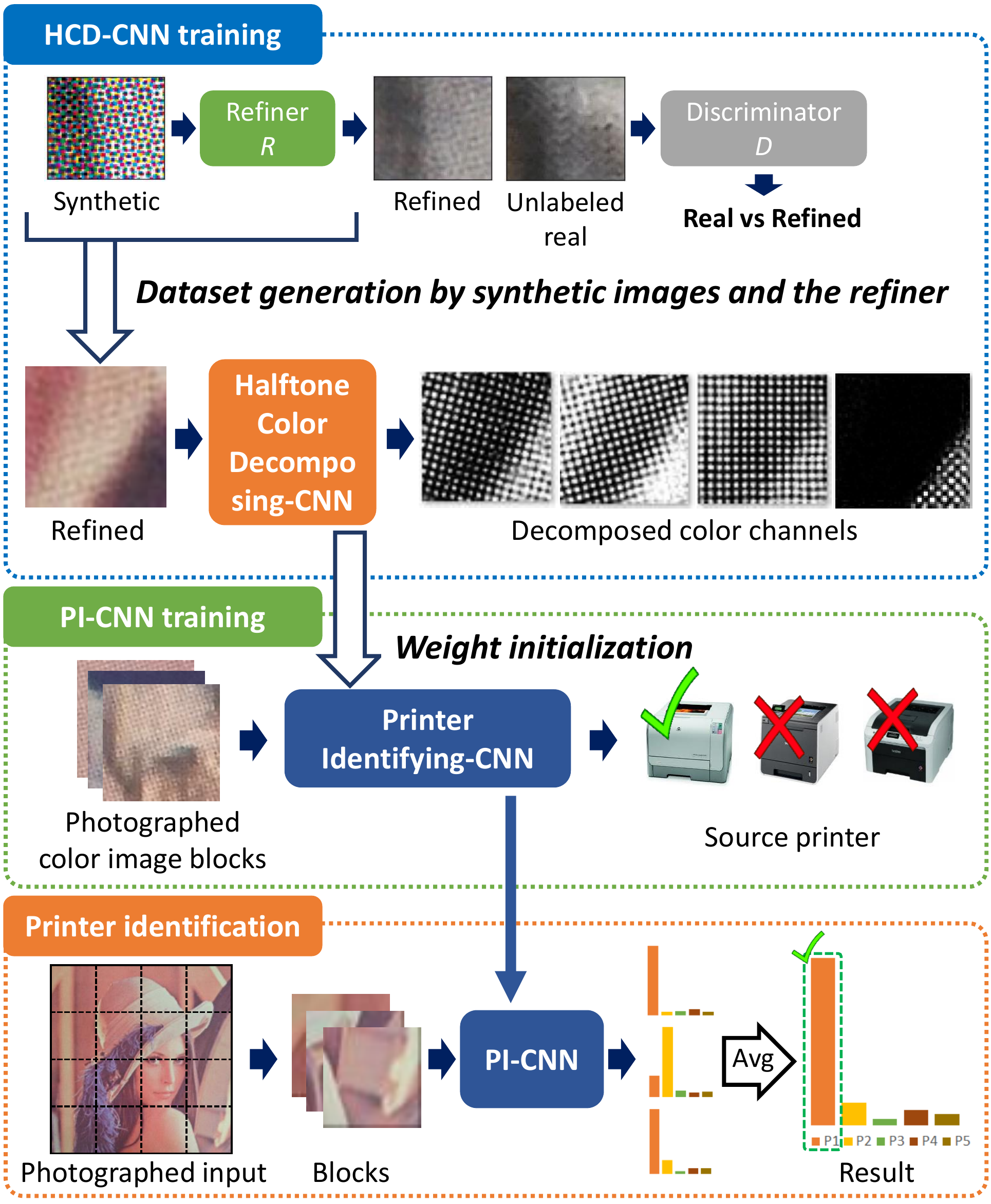}}
    \caption{Overall process of the proposed method
    } \label{Overall}
\end{figure}

\section{Printer identification method}

\subsection{Cascaded learning framework}
As described in Section 2.2, the OPC drum fingerprint is a key feature for the source color laser printer identification. However, it is difficult to extract the OPC drum fingerprint by using existing color decomposition method. Although a machine learning can be used to color decomposition, it is hard to acquire labeled datasets. To resolve these issues, we propose a cascaded learning framework for source color laser printer identification.

The overall process of the proposed method is described in Figure \ref{Overall}. The first step is the HCD-CNN training. We generated the dataset by S+U learning to resolve the issue about the difficulty of acquiring labeled data. The refiner is trained in GAN framework, and the dataset is generated by refining synthetic images. Then, the HCD-CNN is trained with the dataset to resolve the issue about the difficulty of color decomposition. The HCD-CNN decomposes CMYK color channels of given input RGB image. Trained weights of the HCD-CNN are used to initialize weights of the PI-CNN. After that, the PI-CNN is trained with photographed color image blocks. The PI-CNN is designed to decompose halftone components, extract features and classify in a single network. The detailed framework of the PI-CNN is described in Section 3.3. Finally, printer identification is carried out with the trained PI-CNN.

\subsection{Halftone color decomposing-CNN}
\subsubsection{Halftone image refiner training}

Halftone image refiner gets synthesized halftone image as input and refines it to similar with real photographed halftone images. The refined image should have a similar appearance with a real image while preserving the CMYK color channel components from the original synthesized image. The objective of halftone image refiner training is minimizing the following loss:
\begin{equation}
\mathcal{L}_R(\theta)=\sum_{i}\ell_{real}(\theta;x_i,Y)+\lambda\ell_{reg}(\theta;x_i),
\end{equation}
where $x_i$ is the $i^{th}$ image in synthetic images, $Y$ is the real image set, $\theta$ is the parameters of the refiner, and $\lambda$ is a scaling factor. A scaling factor of 10$^{-5}$ is used for the proposed method. The realism loss $\ell_{real}$ is loss about the reality of the refined output, and the self-regularization loss $\ell_{reg}$ is loss about preserving the annotation information about CMYK color channel components.

Following the S+U learning presented in \cite{SimGan}, the refiner $R_\theta$ is trained alternately with the discriminator $D_\phi$, where $\phi$ are the parameters of the discriminator network. The objective of the discriminator network is minimizing the loss:
\begin{equation}
\begin{split}
\mathcal{L}_D(\phi)=-\sum_it_{Fake}\cdot\log(f(D_{\phi}(R_\theta(x_i))))\\-\sum_jt_{Real}\cdot\log(f(D_{\phi}(y_j))),
\end{split}
\end{equation}
where $t$ means label for fake or real, and $f$ means softmax function. The loss is equivalent to cross-entropy error for a two class classification of which label is expressed as a one-hot vector of size 2. The discriminator $D_\phi$ is implemented as a CNN that has two output neurons representing the input is fake or real. The architecture of the discriminator is: (1) Conv3$\times$3, stride=1, feature maps=64, (2) Conv3$\times$3, stride=2, feature maps=64, (3) Conv3$\times$3, stride=1, feature maps=128, (4) Conv3$\times$3, stride=2, feature maps=128, (5) Conv3$\times$3, stride=1, feature maps=256, (6) Conv3$\times$3, stride=2, feature maps=256, (7) FC2. The input is a 64$\times$64 RGB image. The Leaky-ReLU is used for non-linearity function.

Regarding the refiner, we defined the loss $\ell_real$ and $\ell_reg$ as follows:
\begin{equation}
\ell_{real}(\theta;x_i,Y)=-t_{Real}\cdot\log(f(D_{\phi}(R_\theta(x_i)))),
\end{equation}
\begin{equation}
\ell_{reg}(\theta;x_i)=\parallel R_\theta(x_i)-x_i\parallel_2,
\end{equation}
where $\parallel\cdot\parallel_2$ means L2 norm. The refiner is trained to refine input image realistically by minimizing the $\ell_{real}$ loss. The $\ell_{reg}$ is needed to preserve the annotation information of the synthetic images. The refiner $R_\theta$ is implemented as a fully convolutional neural net (FCN). We adopted refined image history buffer suggested in \cite{SimGan} to stabilize the refiner training. The architecture of the refiner is: (1) Conv3$\times$3, stride=1, feature maps=64, (2) Conv3$\times$3, stride=1, feature maps=64, (3) Conv3$\times$3, stride=1, feature maps=64, (4) Conv3$\times$3, stride=2, feature maps=64, (5) Conv3$\times$3, stride=1, feature maps=64, (6) Conv3$\times$3, stride=2, feature maps=64, (7) Conv3$\times$3, stride=2, feature maps=16, (8) Conv3$\times$3, stride=2, feature maps=4. The input is a 64$\times$64 RGB image. The ReLU is used for non-linearity function except for the last layer that used Tanh. The detail process of the refiner training is described in Algorithm 1.

\begin{algorithm}[t]
	\caption{Halftone image refiner training process}
	\begin{algorithmic}[1]
		\INPUT image buffer $B$, mini-batch size $b$, sets of synthetic images $x_i \in X$, and real images $y_j \in Y$, max iteration number $T$
		\OUTPUT FCN model $R_\theta$
		\STATE Set $B$ as a empty buffer
		\FOR{$t=1,...,T$}
			\FOR{$k = 1,2$}
			\STATE Sample a mini-batch input of $x_i$
			\IF{$B$ is not a maximum size}
				\STATE Append $R_\theta(x_i)$ to the $B$
			\ELSE
				\STATE Replace $b/2$ images in the $B$ with $R_\theta(x_i)$
			\ENDIF
			\STATE Update $\theta$ by taking a SGD step on the $\mathcal{L}_R(\theta)$ calculated from the mini-batch
			\ENDFOR
		\STATE Sample a mini-batch input of $x_i$ and $y_j$
		\STATE Sample $b/2$ images in the $B$ and $b/2$ images $R_\theta(x_i)$ with current $\theta$ and merge images to create refined input of the discriminator
		\STATE Update $\phi$ by taking a SGD step on the $\mathcal{L}_D(\phi)$ calculated from the mini-batch
		\ENDFOR
	\end{algorithmic}
\end{algorithm}

\subsubsection{Halftone color decomposing-CNN training}

After the refiner training is completed, we refined all synthetic images to prepare the dataset for HCD-CNN training. Then, the HCD-CNN is trained with the refined dataset. The objective of the HCD-CNN is decomposing CMYK toner information that is mixed in RGD color domain. The trained HCD-CNN is used for transferring the knowledge about halftone color decomposition to the PI-CNN.

The architecture of the HCD-CNN is: (1) Conv3$\times$3, stride=1, feature maps=64, (2) Conv3$\times$3, stride=1, feature maps=64, (3) Conv3$\times$3, stride=1, feature maps=64, (4) Conv3$\times$3, stride=1, feature maps=64, (5) Conv3$\times$3, stride=1, feature maps=64, (6) Conv3$\times$3, stride=1, feature maps=64, (7) Conv3$\times$3, stride=1, feature maps=64, (8) Conv3$\times$3, stride=1, feature maps=4, (9) Euclidean loss. The input is a 64$\times$64 RGB image. The ReLU is used for non-linearity function, and batch normalization is applied to every convolutional layer. Each feature map of the output means decomposed CMYK color channel of the input image, respectively.

\begin{figure}[t!]
    \centerline{\includegraphics[width=9cm]{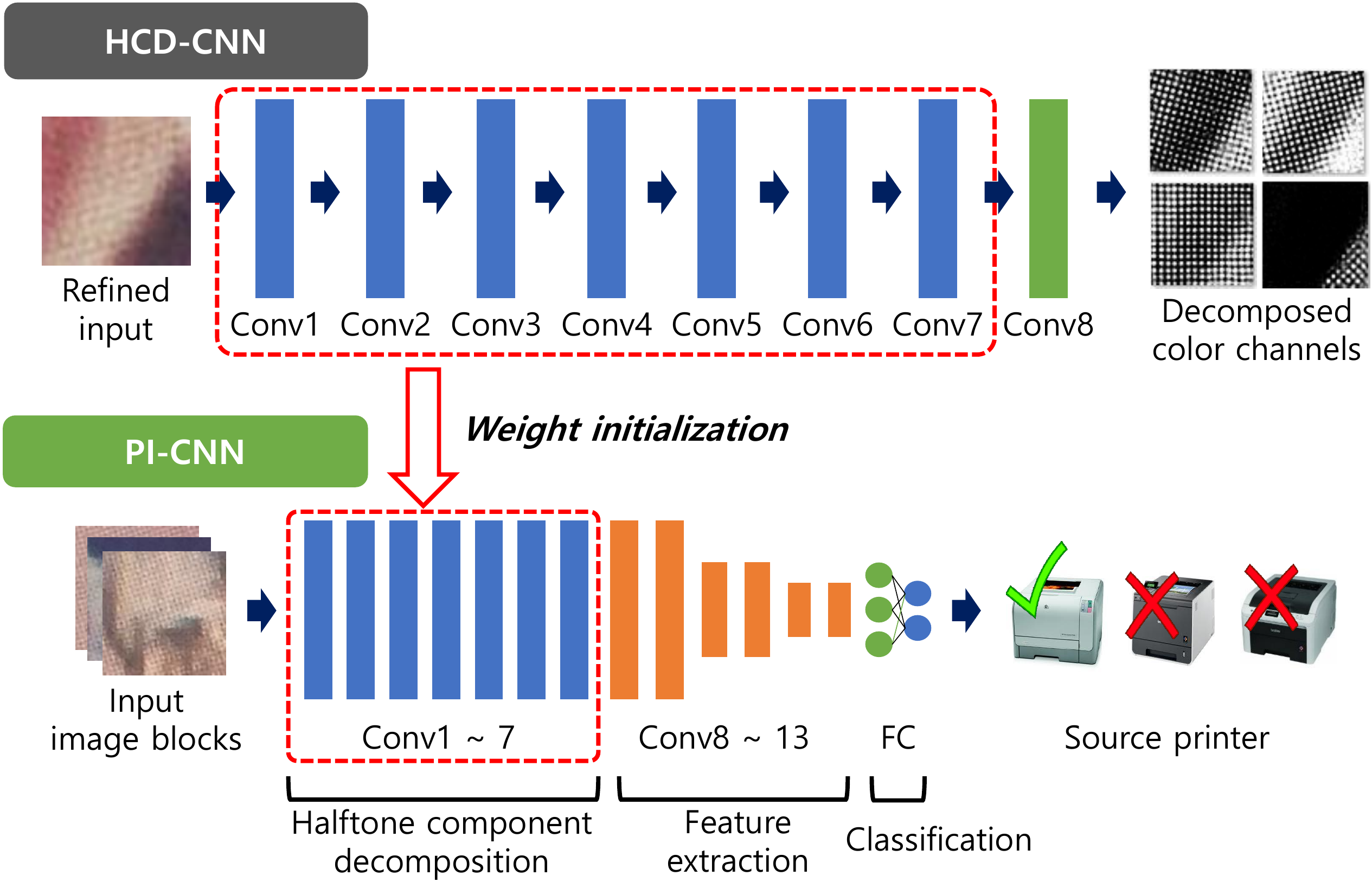}}
    \caption{Parameter transferring between the HCD-CNN and the PI-CNN
    } \label{Transfer}
\end{figure}

\begin{figure}[t]
    \centerline{\includegraphics[width=8.5cm]{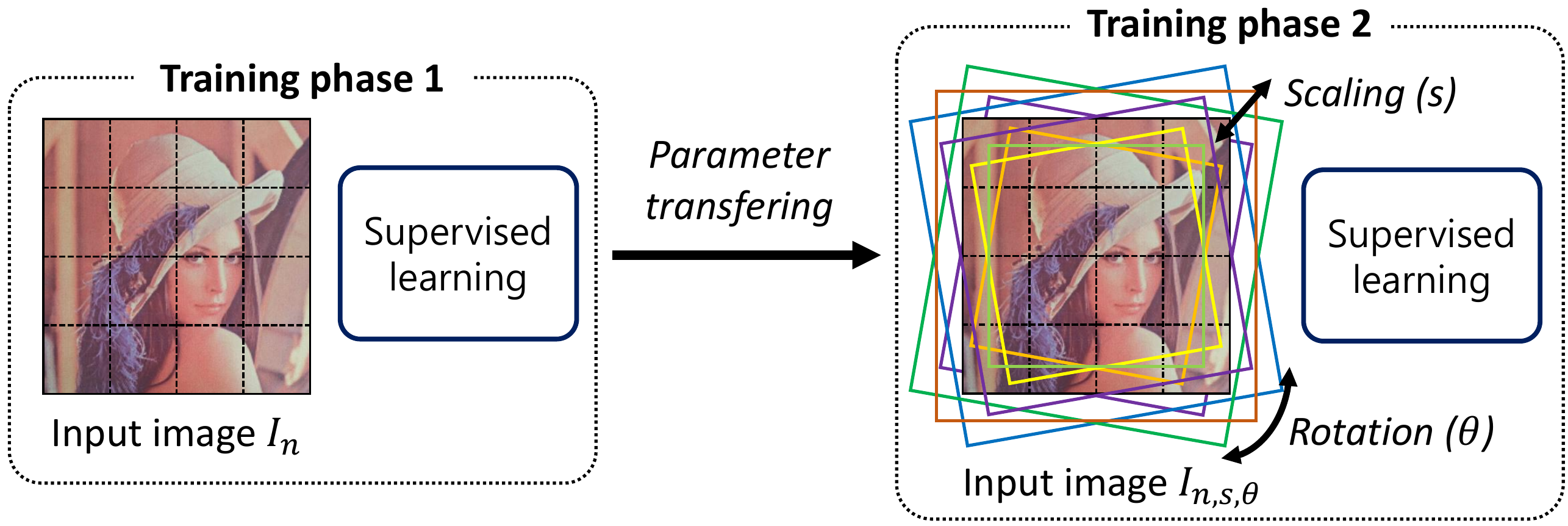}}
    \caption{PI-CNN training process
    } \label{Robust}
\end{figure}

\subsection{Printer identifying-CNN}

The objective of the PI-CNN is identifying the source color laser printer of input RGB image block. To utilize the knowledge about halftone color decomposition of the HCD-CNN, part of the weights of the PI-CNN is initialized with the trained weights of the HCD-CNN as presented in Figure \ref{Transfer}. As shown in Figure \ref{Transfer}, the PI-CNN is composed of three parts that carry out halftone component decomposition, feature extraction, classification respectively. Halftone component decomposition part is initialized with the weights of the HCD-CNN, and other parts are initialized with Xavier initialization \cite{xavier}. Then, all layers of the PI-CNN are trained in the gradient descent process.

The training process is composed of two phases as shown in Figure \ref{Robust} to achieve robustness about input scaling and rotation. In the first phase, the PI-CNN is trained with the photographed color document input $I_n$. $I_n$ isn't scaled and rotated. Next, weights of the trained PI-CNN are transferred and fine-tuned to achieve robustness about scaling and rotation. Input image $I_{(n,s,\theta)}$ has scaling factor $s$ and rotation factor $\theta$ that are randomly selected as $s\in\{0.8,1.0,1.2\},\theta\in\{-10^\circ,0^\circ,+10^\circ\}$.

The architecture of the PI-CNN used in the proposed method is follows: (1) Conv3$\times$3, stride=1, feature maps=64, (2) Conv3 $\times$3, stride=1, feature maps=64, (3) Conv3$\times$3, stride=1, feature maps=64, (4) Conv3$\times$3, stride=1, feature maps=64, (5) Conv3$\times$3, stride =1, feature maps=64, (6) Conv3$\times$3, stride=1, feature maps=64, (7) Conv3$\times$3, stride=1, feature maps=64, (8) Conv3$\times$3, stride=1, feature maps=64, (9) Conv3$\times$3, stride=1, feature maps=64, (10) MaxPool2$\times$2, stride=2, (11) Conv3$\times$3, stride=1, feature maps=128, (12) Conv3$\times$3, stride=1, feature maps=128, (13) MaxPool2$\times$2, stride=2, (14) Conv3$\times$3, stride= 1, feature maps=256, (15) Conv3$\times$3, stride=1, feature maps= 256, (16) MaxPool2$\times$2, stride=2, (17) FC4096, (18) FC4096, (19) FC8, (20) Softmax. The input is a 64$\times$64 RGB image, and the output is a vector consisting of eight neuron values, which denotes the number of the source printers used in the experiment. If the number of the candidate source printers is changed, the number of neurons in the last fully-connected layer can be changed. The ReLU is used for non-linearity function and batch normalization is applied to every convolutional layer.

\begin{figure}[t]
    \centerline{\includegraphics[width=8.5cm]{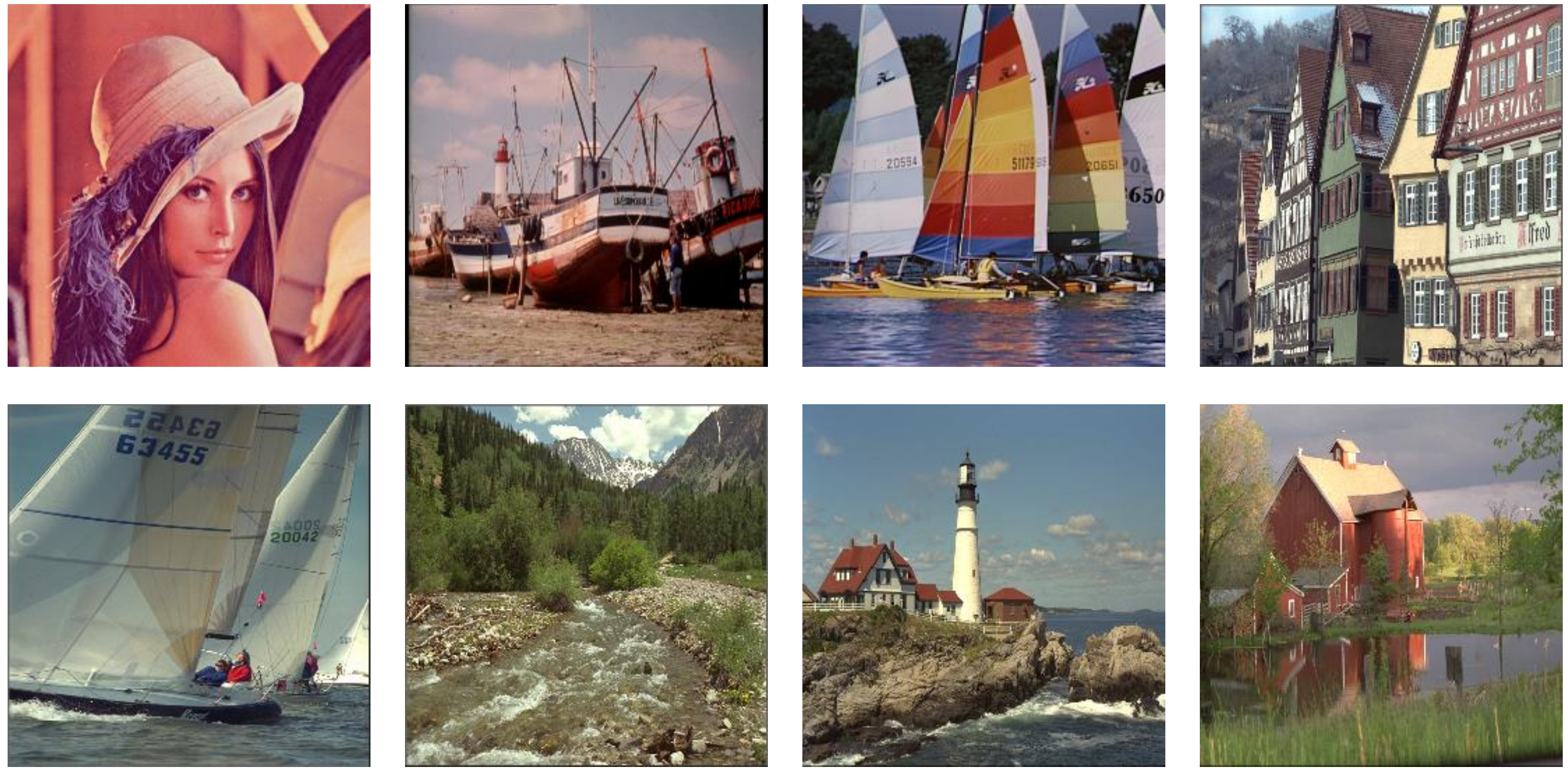}}
    \caption{Example images used in experiments
    } \label{ExpImgs}
\end{figure}

\subsection{Source color laser printer identification}

The trained PI-CNN gets image blocks as inputs. A photographed color document image is much bigger than the PI-CNN input size. Therefore we divide the image into blocks and merge all softmax result of block feed-forwardings to the PI-CNN. The source color laser printer is identified based on the average of softmax outputs as the following equation:

\begin{equation}
p_s = \argmax_{i=1\sim n} m^{-1}\sum^{m}_{j=1}f_j(z_i),
\end{equation}
where $p_s$ denotes source printer, $n$ is the number of candidate source printers, $m$ is the number of input blocks, $f_j(z_i)$ denotes softmax output of $i$-th candidate printer where the input block is $j$-th block.

\begin{figure*}[t!]
    \centerline{\includegraphics[width=12cm]{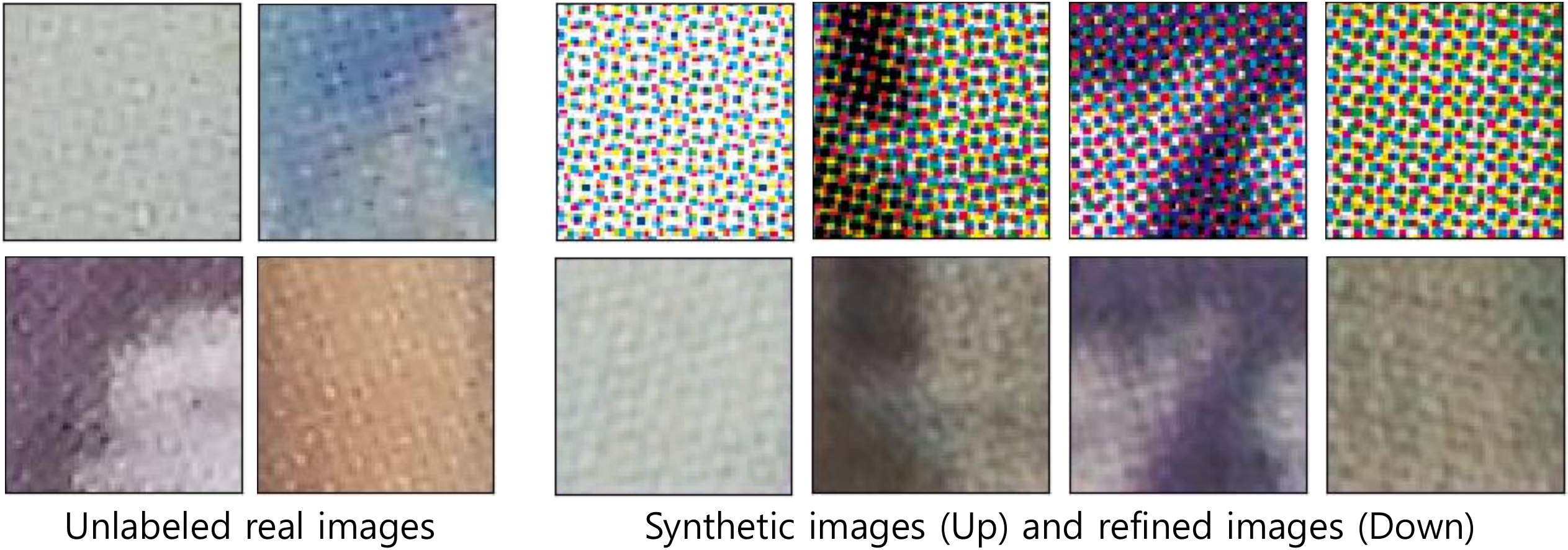}}
    \caption{Example output of the trained refiner
    } \label{Refined}
\end{figure*}

\begin{figure*}[t!]
    \centerline{\includegraphics[width=10cm]{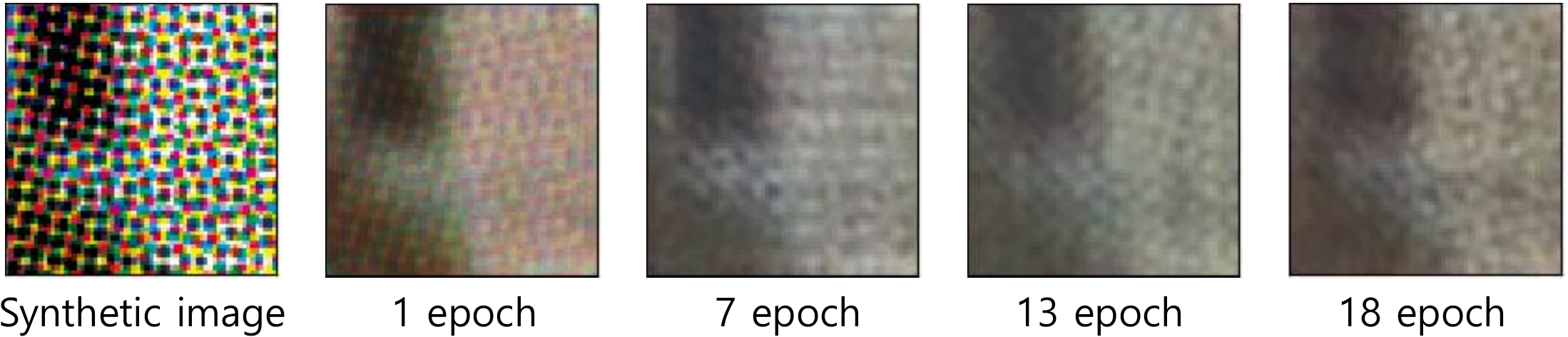}}
    \caption{Example output of the refiner during the training
    } \label{RefineEpoch}
\end{figure*}

\section{Experimental results and discussion}
\subsection{Experimental environment}
\subsubsection{Refiner and HCD-CNN}

Example images used in our experiments are shown in Figure \ref{ExpImgs}. Same images were used in all steps of the proposed method. In the refiner training, images halftoned with Adobe Photoshop CS 6 were used as synthetic images. The ratio between halftone dot size and image size is set to equal with the ratio between toner dot size and printed image size. The images printed from eight color laser printers and photographed by using a Galaxy Note 3 (Samsung) smartphone were used as unlabeled real images. Source color laser printers are listed in Table \ref{printers}. The images were also used for printer identification.

In all training process of neural networks, we used Adam optimizer. The refiner was trained with 10$^{-5}$ learning rate and 32 batch size. The HCD-CNN was trained with 10$^{-4}$ learning rate and 32 batch size. Training was stopped when the validation loss converges. All neural networks were implemented using the TensorFlow \cite{tf} library.

\begin{table}[t]
\small
\caption{A list of printers used in experiments}
\begin{center}
\begin{tabular}{lcc}
\hline
\textbf{Label}    & \textbf{Brand} & \textbf{Model} \\
\hline
H1  & HP & HP 4650 \\
H2 & HP & HP CM3530 \\
X1 & Xerox & 700 Digital Color Press \\
X2 & Xerox & Docu Centre C450 \\
X3 & Xerox & Docu Centre C6500 \\
K1 & Konica Minolta & Bizhub Press C280 \\
K2 & Konica Minolta & Bizhub Press C280 \\ 
K3 & Konica Minolta & Bizhub Press C8000 \\
\hline
\end{tabular}
\end{center}
\label{printers}
\end{table}

\subsubsection{Printer identification}
A total of eight printers from three brands listed in Table \ref{printers} were used for the experiments. The proposed method, Kim's method using halftone texture fingerprints \cite{Kim2}, Kim's method using close-up photography \cite{Kim1}, Tsai's method \cite{Tsai1}, and Ryu's method \cite{Ryu} were used in the experiment to compare the performances. As Tsai's hybrid method \cite{Tsai2} used not only printed images but also printed texts, it was not used in the comparison experiment. Choi's methods \cite{Choi1}\cite{Choi2} were not used in the experiment because Tsai's method \cite{Tsai1} was based on Choi's methods, and it had slightly better performance than Choi's methods.

The images that used in refiner and HCD-CNN training were used for experiments. The size of original images photographed with the smartphone was 2322$\times$4128. The photographing distance and the photographing angle were kept equal in all input images. We cropped original images, and a total of 768 images from each color laser printer was used for the experiment. The size of input images was 512$\times$512. 2-fold cross-validation was adopted for the test of the existing methods. In the experiment of the proposed method, 49,152 image blocks for each color laser printer (extracted from same input images used for existing methods) were used for the training. 2-fold cross-validation is also adopted for the test of the proposed method in a modified format. To use early stopping, we divided validation set of cross-validation into two sets. Each set was used as validation set and test set alternatively. Thus, there were four test results for the proposed method. Data augmentation was not used in the experiment. The PI-CNN was trained with 2$\times10^{-5}$ learning rate and 32 batch size. The training was stopped when the validation accuracy is not increased for ten epochs.

\begin{table}
\small
\caption{Refined halftone image decomposition results}
\begin{center}
\begin{tabular}{ l |*{4}{c|}}
\hline \multicolumn{1}{ |c|}{\multirow{2}{*}{Channel}} & \multicolumn{2}{c|}{PSNR (dB)} & \multicolumn{2}{c|}{SSIM}\\
\cline{2-5} \multicolumn{1}{|c|}{} & HCD-CNN & Profile & HCD-CNN & Profile \\
\hline \multicolumn{1}{|c|}{Cyan} & \textbf{17.8953} & 5.7045 & \textbf{0.8412} & 0.0093\\
\hline \multicolumn{1}{|c|}{Magenta} & \textbf{17.8108} & 5.8335 & \textbf{0.8172} & 0.0094\\
\hline \multicolumn{1}{|c|}{Yellow} & \textbf{18.5375} & 6.6622 & \textbf{0.7537} & 0.0245\\
\hline \multicolumn{1}{|c|}{Black} & 26.5491 & \textbf{29.5980} & \textbf{0.7725} & 0.5192\\
\hline
\end{tabular}
\end{center}
\label{hcd_table}
\end{table}

\begin{figure}[t]
    \centerline{\includegraphics[width=8.8cm]{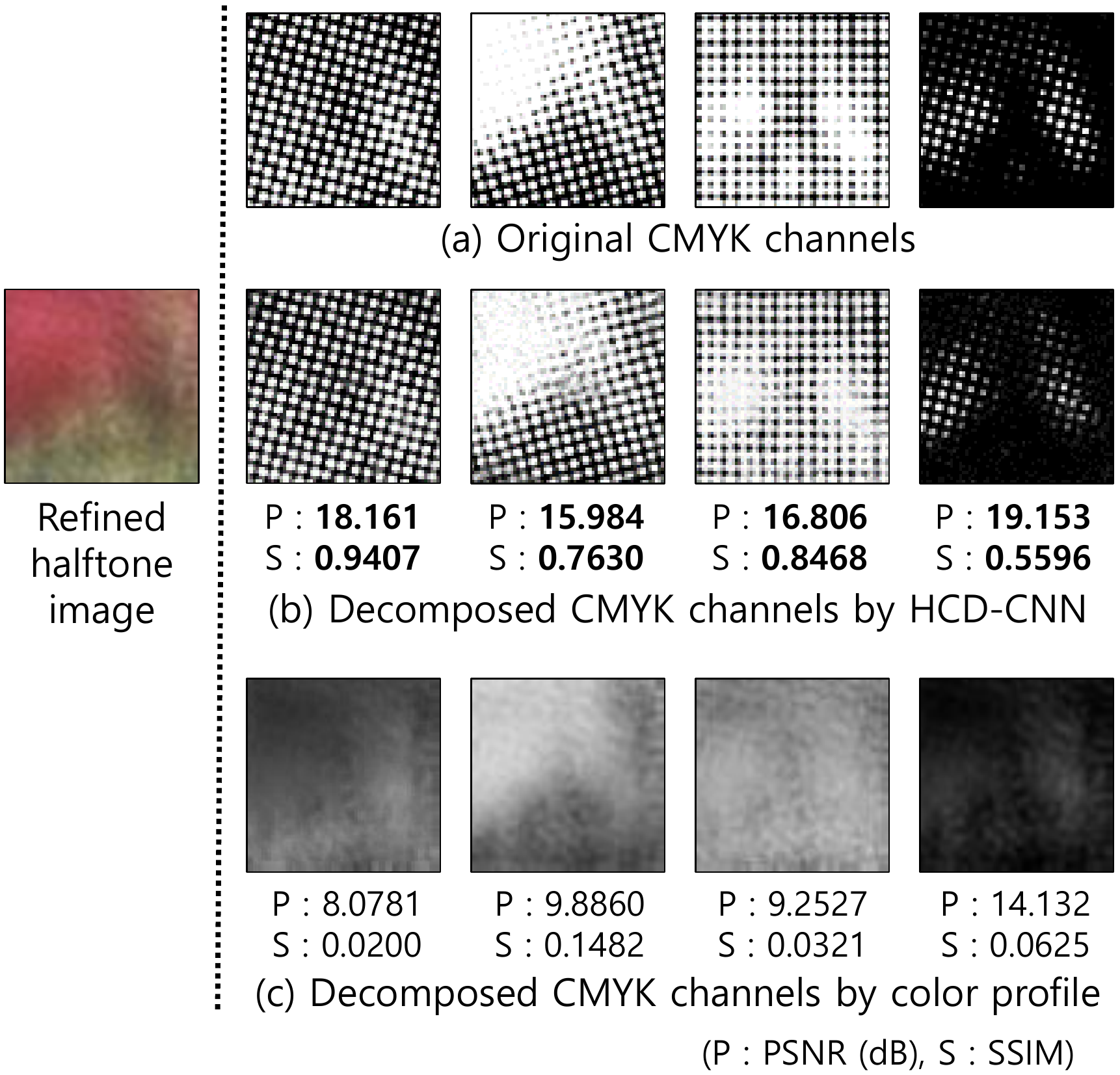}}
    \caption{Halftone color decomposition of refined image
    } \label{HCD_refine}
\end{figure}

\subsection{Refiner training results}
Figure \ref{Refined} shows examples of synthetic, real and refined color halftone images. As shown in example images, synthetic images show clear halftone dots while real photographed halftone images show blurred halftone dots and mixed patterns. Refined images show blurred halftone dots and patterns that are similar to real images. The trained refiner successfully refine synthetic images to be looked as real while preserving annotation information.

Example outputs of the refiner at various training epochs are shown in Figure \ref{RefineEpoch}. The refiner produced unrealistic artifacts when it trained for just one epoch. During the training, the refiner learned to refine synthetic images to be looked as real. It produced realistic images after trained ten epochs. We used the refiner trained for 18 epochs since the validation loss was converged from that point.

\begin{figure}[t]
    \centerline{\includegraphics[width=8.8cm]{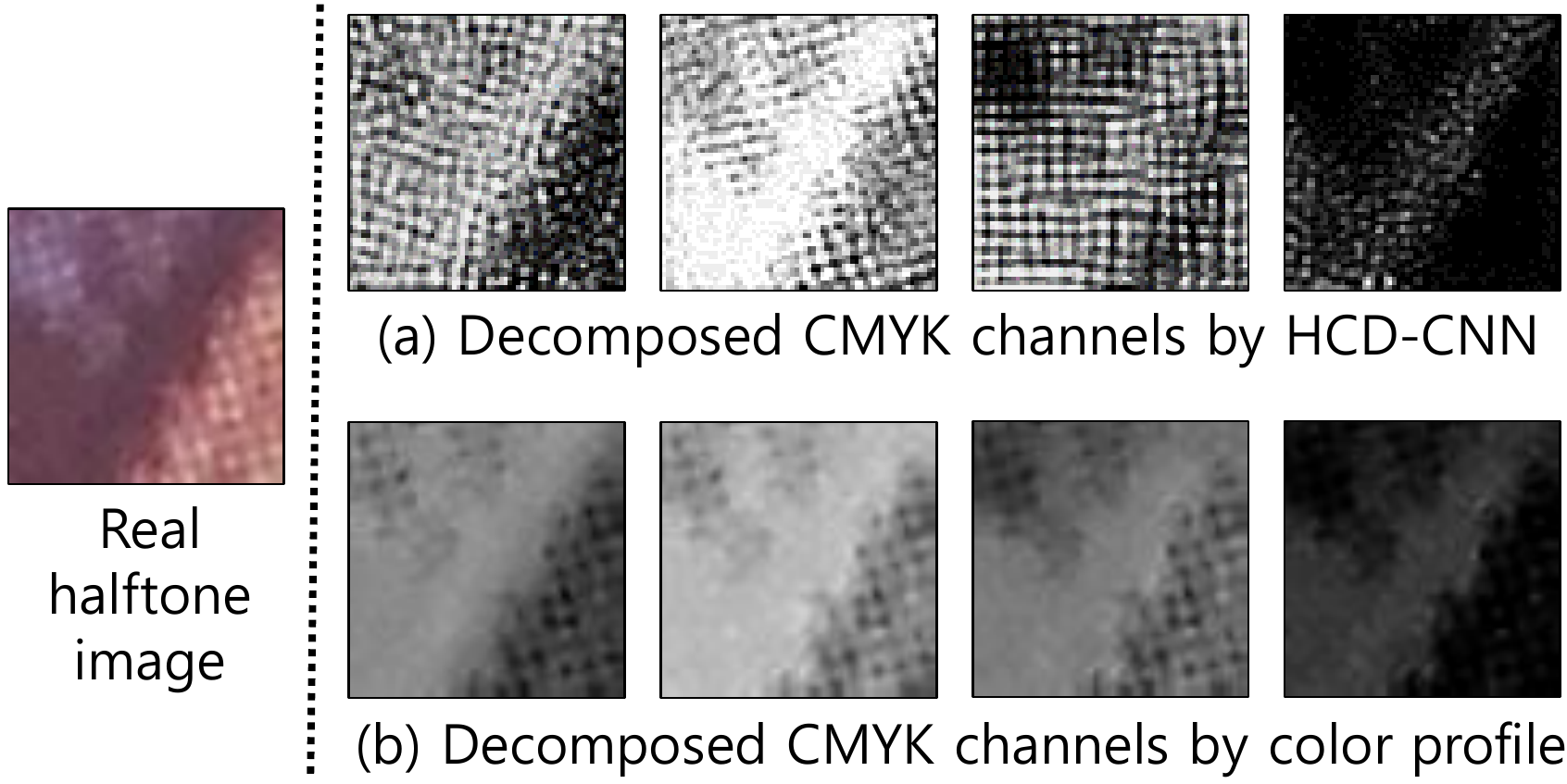}}
    \caption{Halftone color decomposition of real image
    } \label{HCD_real}
\end{figure}

\subsection{HCD-CNN training results}
Performance comparison results between the HCD-CNN and existing halftone color decomposition using pre-defined color profile are presented in Figure \ref{HCD_refine} and Table \ref{hcd_table}. Figure \ref{HCD_refine} (a) is original CMYK channels of a refined halftone image not used for training, Figure \ref{HCD_refine} (b) is decomposed CMYK channels by using HCD-CNN, and Figure \ref{HCD_refine} (c) is decomposed CMYK channels by existing method using color profile; U.S. Sheetfed Coated profile was used to convert color domain. Table \ref{hcd_table} presents peak signal-to-noise ratio (PSNR) and structural similarity (SSIM) between the decomposed color channel and original color channel of 12,288 refined halftone images not used for training. The higher PSNR and SSIM mean that the decomposed color channel is similar to the original pattern.

As shown in Figure \ref{HCD_refine}, the HCD-CNN decomposed each CMYK channels from blurred and mixed halftone patterns. The existing method couldn't separate mixed halftone dots so that decomposed channels are blurred. The HCD-CNN shows better performances in all measures of Figure \ref{HCD_refine}. In the result of Table \ref{hcd_table}, all measures of the HCD-CNN is better than the existing method except for the PSNR of the black channel. Notably, the performance gap in SSIM of CMY channels is much higher than other measures.

\begin{figure}[t!]
    \centerline
    {
    	\subfigure[Phase 1] {\includegraphics[width=4.3cm]{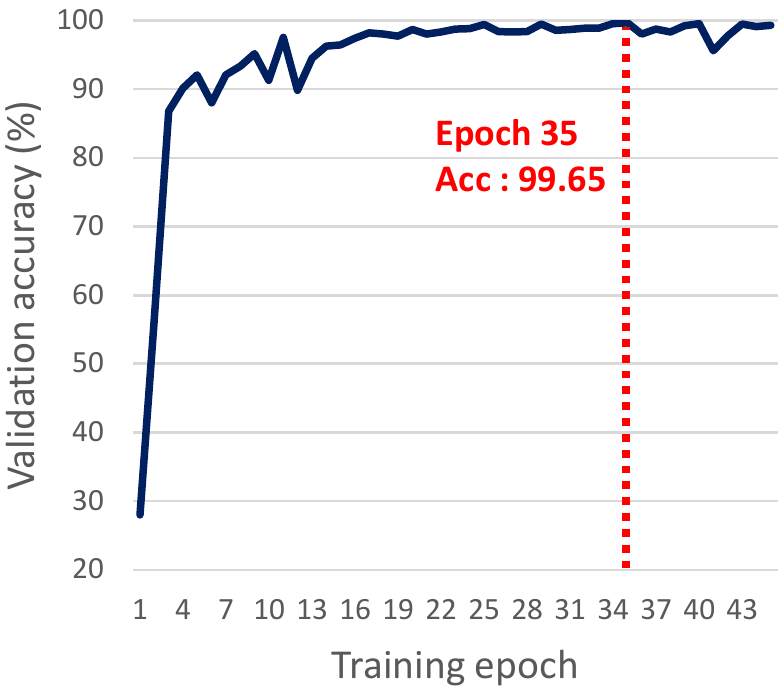}	\label{Phase1}}
    	    	\hspace{2pt}
	\subfigure[Phase2]	{\includegraphics[width=4.3cm]{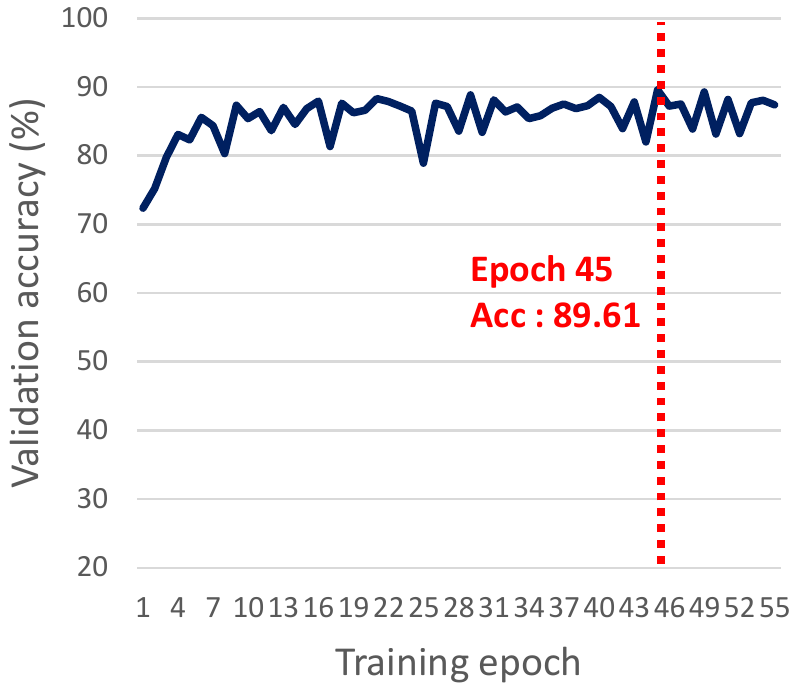}	\label{Phase2}}
    }
    \caption{Validation accuracy during the training}
    \label{Phase1_2}	
\end{figure}

\begin{figure}[t]
    \centerline{\includegraphics[width=8.8cm]{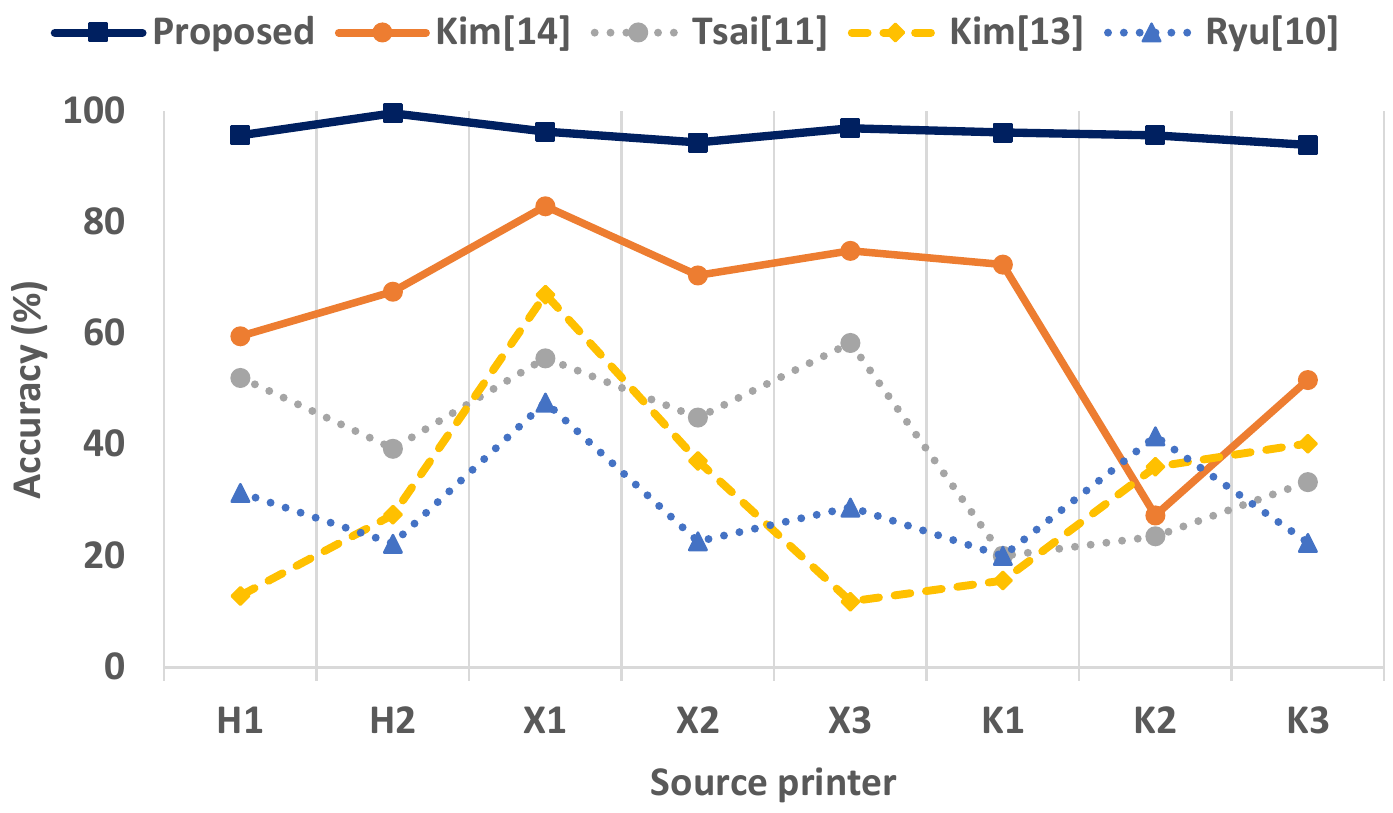}}
    \caption{Printer identification results graph
    } \label{Result}
\end{figure}

\begin{table*}[t]
\small
\caption{Printer identification results}
\begin{center}
\begin{tabular}{ l |*{9}{c|}}
\hline \multicolumn{1}{ |c|}{\multirow{2}{*}{Methods}} & \multicolumn{9}{c|}{Identification accuracy (\%)} \\
\cline{2-10} \multicolumn{1}{|c|}{} & H1 & H2 & X1 & X2 & X3 & K1 & K2 & K3 & Avg.$\pm$ Std.Dev \\
\hline \multicolumn{1}{|c|}{Kim\cite{Kim2}} & 59.51 & 67.58 & 82.94 & 70.44 & 74.87 & 72.40 & 27.34 & 51.69 & 63.35$\pm$1.60\\
\hline \multicolumn{1}{|c|}{Tsai\cite{Tsai1}} & 52.08 & 39.32 & 55.60 & 44.92 & 58.33 & 20.05 & 23.57 & 33.33 & 40.90$\pm$1.42\\
\hline \multicolumn{1}{|c|}{Kim\cite{Kim1}} & 12.89 & 27.47 & 67.06 & 37.11 & 11.85 & 15.63 & 36.07 & 40.23 & 31.04$\pm$0.44\\
\hline \multicolumn{1}{|c|}{Ryu\cite{Ryu}} & 31.38 & 22.27 & 47.66 & 22.66 & 28.78 & 20.05 & 41.54 & 22.40 & 29.59$\pm$1.50\\
\hline \multicolumn{1}{|c|}{Proposed} & \textbf{95.67} & \textbf{99.64} & \textbf{96.29} & \textbf{94.40} & \textbf{96.94} & \textbf{96.19} & \textbf{95.61} & \textbf{93.95} & \textbf{96.09$\pm$2.37}\\
\hline
\end{tabular}
\end{center}
\label{result_table}
\end{table*}

\begin{figure*}[t!]
    \centerline
    {
    	\subfigure[Proposed] {\includegraphics[width=4.5cm]{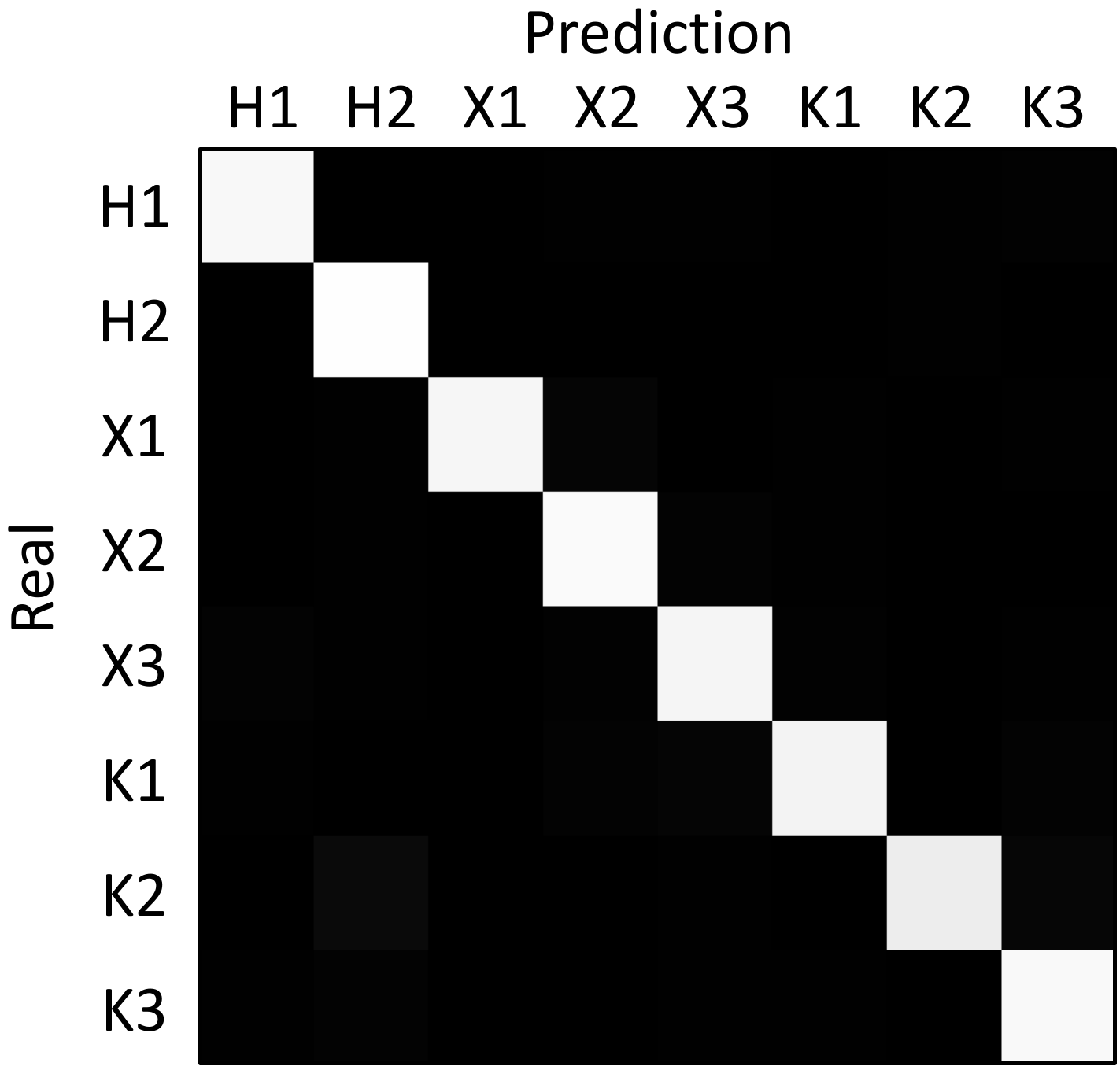}	\label{ProposedConf}}
    	    	\hspace{2pt}
	\subfigure[Kim\cite{Kim2}]	{\includegraphics[width=4.5cm]{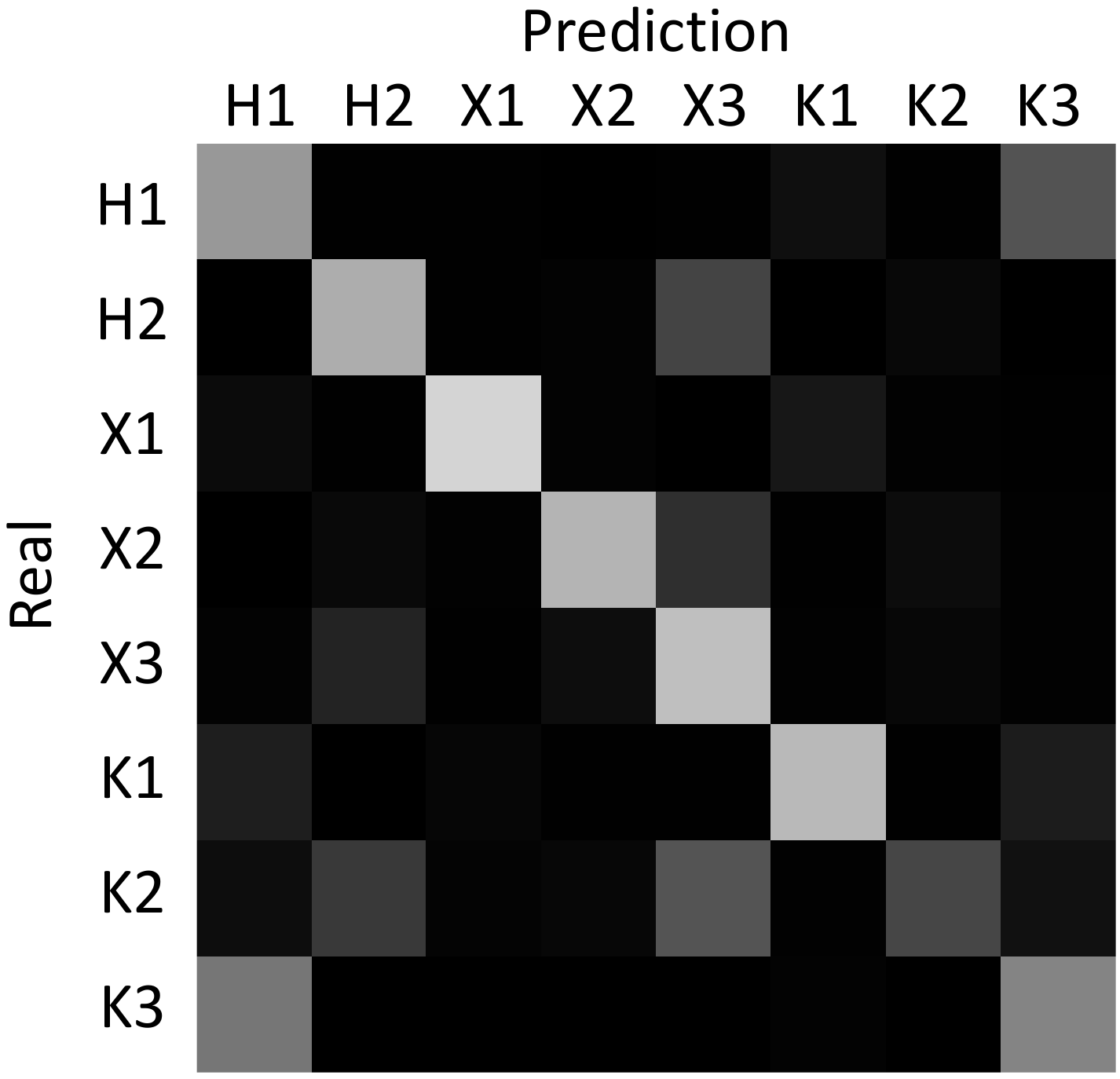}	\label{Kim2Conf}}
	\subfigure[Tsai\cite{Tsai1}]	{\includegraphics[width=4.5cm]{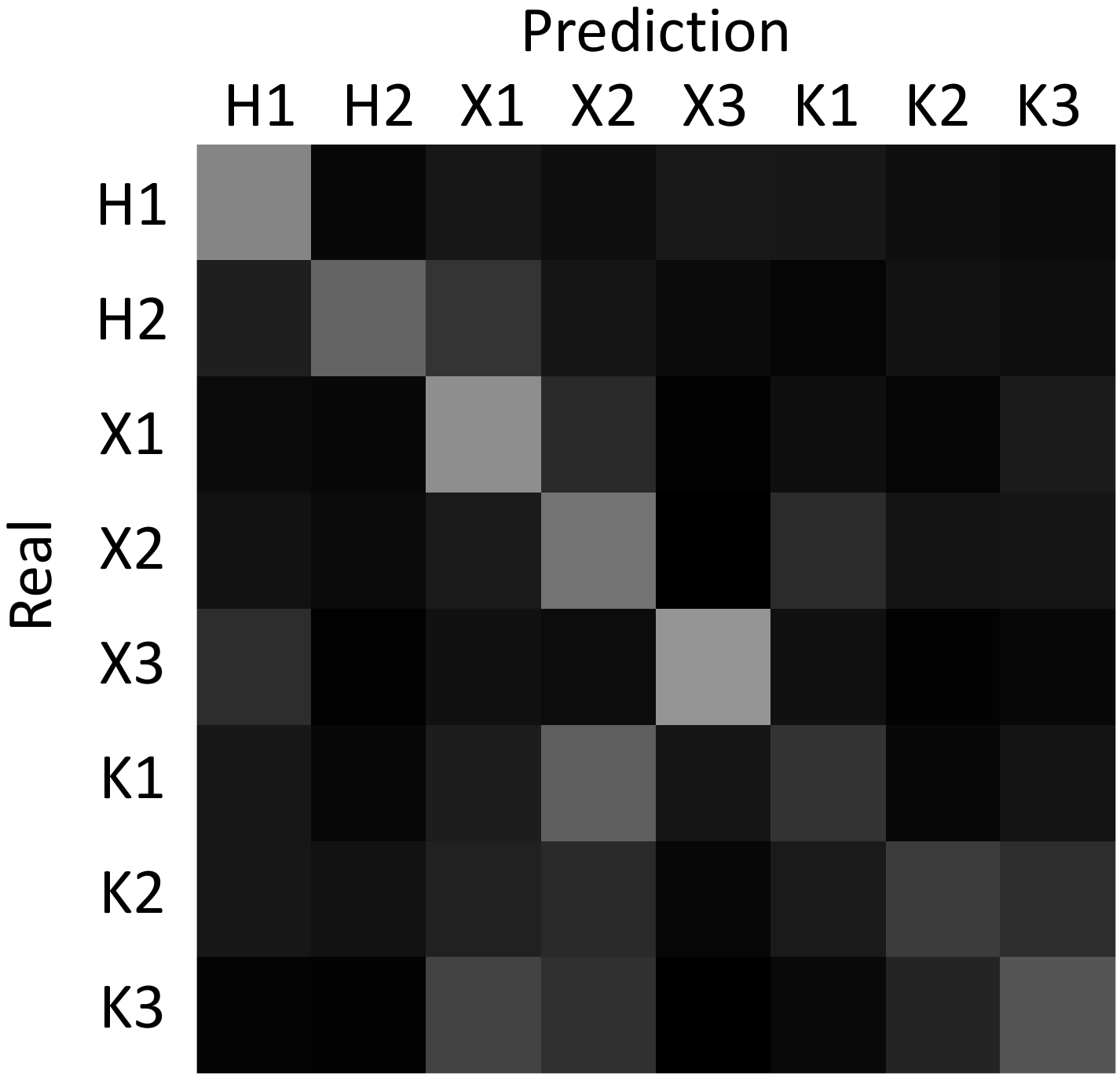}	\label{TsaiConf}}
    }
	\centerline{	
    	\subfigure[Kim\cite{Kim1}]	{\includegraphics[width=4.5cm]{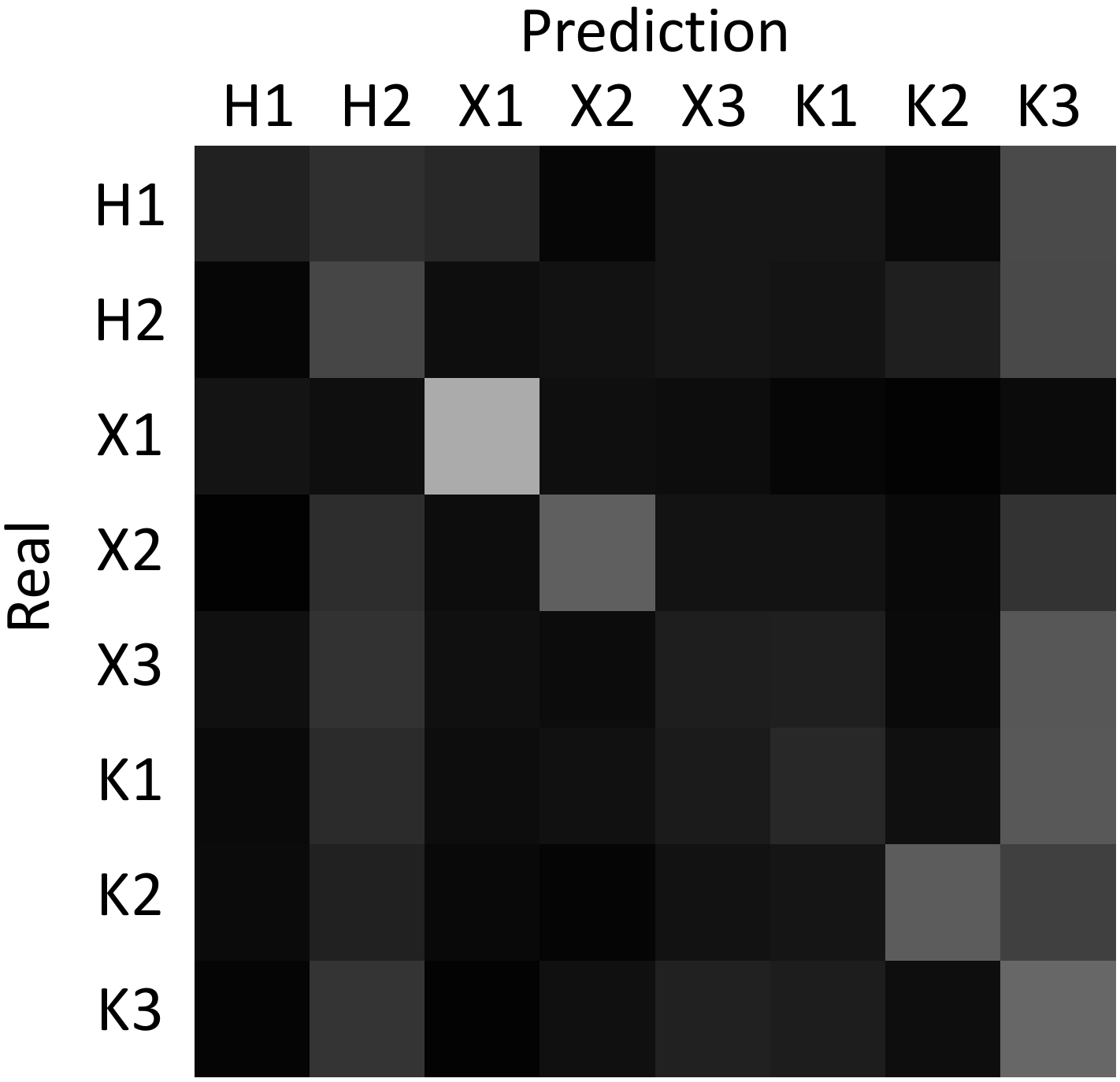}	\label{Kim1Conf}}
    	    	\hspace{2pt}
	\subfigure[Ryu\cite{Ryu}]	{\includegraphics[width=4.5cm]{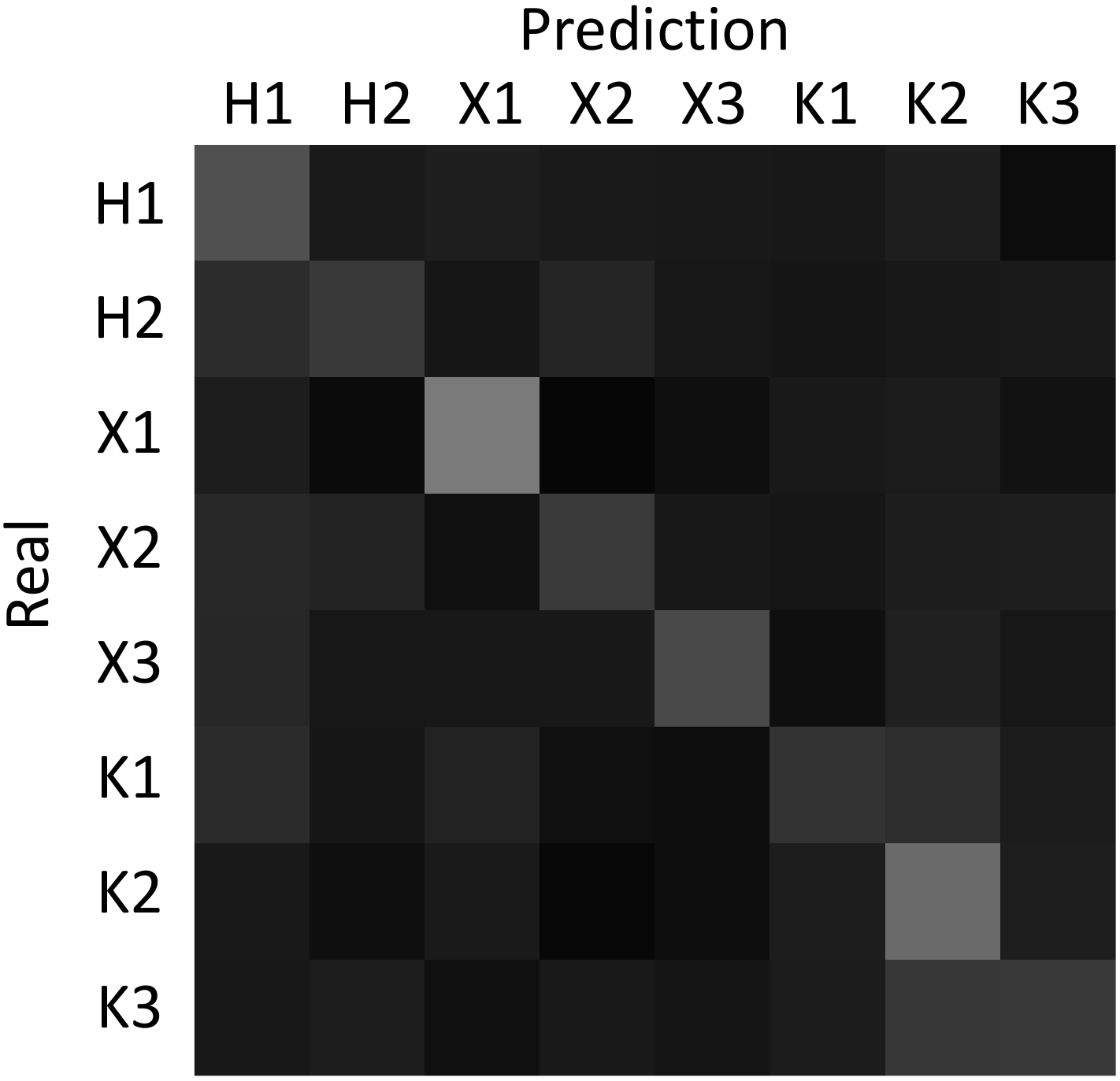}	\label{RyuConf}}
	}
    \caption{Graphical confusion matrices}
    \label{Confusion}	
\end{figure*}

In Figure \ref{HCD_real}, results of halftone color decomposition of the real image are shown. In comparison with the existing method, the HCD-CNN suggested halftone dots of each CMYK channel while existing method just decomposes each CMYK channels as the same pattern with different local intensity. The knowledge about decomposing halftone color channels of the HCD-CNN was transferred to the PI-CNN, and the PI-CNN showed overwhelming source printer identification accuracy compared to existing methods that are using color profile based halftone color decomposition.

\subsection{PI-CNN training results}
The validation accuracy during the PI-CNN training is presented in Fig. \ref{Phase1_2}. Among all four cross-validation test results, Fig. \ref{Phase1_2} represents one of them. The validation accuracy during the training phase 1 and the training phase 2 is described in Fig. \ref{Phase1} and Fig. \ref{Phase2}, respectively. As shown in Fig. \ref{Phase1}, the validation accuracy did not increase for ten epochs after epoch 35. Therefore, weights of the network at epoch 35 were used in the training phase 2. In the training phase 2, weights of the network at epoch 45 were used for the test based on the early stopping rule. Average identification accuracy of the trained networks at the blockwise test was 88.37\%, and the standard deviation was 1.25.

\subsection{Printer identification results}
\subsubsection{Identification accuracy evaluation}
The printer identification results are summarized in Fig. \ref{Result} and Table \ref{result_table}. The average accuracy of the proposed method was 96.09\%, which is the highest accuracy among the all five tested methods. Kim's method using halftone fingerprints had a better performance than the other three existing methods, but its average accuracy was lower by about 33\% than the proposed method. It was hard for the other methods to identify the source color laser printers because the input images were photographed images while these methods were designed to identify the source printer of close-up photographed images \cite{Kim1} or scanned images \cite{Ryu}\cite{Tsai1}.

\begin{table*}[t]
\small
\caption{Robustness evaluation results for rotation (Unit: \%)}
\begin{center}
\begin{tabular}{ l |*{5}{c|}}
\hline \multicolumn{1}{ |c|}{\multirow{2}{*}{Methods}} & \multicolumn{5}{c|}{Rotation degree ($^\circ$), Accuracy (Avg.$\pm$ Std.Dev)} \\
\cline{2-6} \multicolumn{1}{|c|}{} & $-$10 & $-$5 & 0 & $+$5 & $+$10 \\
\hline \multicolumn{1}{|c|}{Kim\cite{Kim2}} & 16.63$\pm$0.65 & 28.71$\pm$0.62 & 63.35$\pm$1.60 & 31.22$\pm$0.36 & 14.55$\pm$0.23\\
\hline \multicolumn{1}{|c|}{Tsai\cite{Tsai1}} & 13.64$\pm$0.85 & 13.83$\pm$1.17 & 40.90$\pm$1.42 & 13.67$\pm$1.04 & 13.49$\pm$0.76\\
\hline \multicolumn{1}{|c|}{Kim\cite{Kim1}} & 17.92$\pm$1.97 & 18.60$\pm$1.17 & 31.04$\pm$0.44 & 21.74$\pm$0.65 & 22.04$\pm$1.14\\
\hline \multicolumn{1}{|c|}{Ryu\cite{Ryu}} & 11.62$\pm$0.68 & 19.01$\pm$0.03 & 29.59$\pm$1.50 & 13.79$\pm$0.24 & 10.68$\pm$0.20\\
\hline \multicolumn{1}{|c|}{Proposed} & \textbf{95.00$\pm$2.49} & \textbf{93.01$\pm$2.98} & \textbf{96.09$\pm$2.37} & \textbf{92.80$\pm$2.97} & \textbf{94.46$\pm$2.47}\\
\hline
\end{tabular}
\end{center}
\label{Rotation_table}
\end{table*}

\begin{table*}[t]
\small
\caption{Robustness evaluation results for scaling (Unit: \%)}
\begin{center}
\begin{tabular}{ l |*{5}{c|}}
\hline \multicolumn{1}{ |c|}{\multirow{2}{*}{Methods}} & \multicolumn{5}{c|}{Scaling factor, Accuracy (Avg.$\pm$ Std.Dev)} \\
\cline{2-6} \multicolumn{1}{|c|}{} & 0.8 & 0.9 & 1.0 & 1.1 & 1.2 \\
\hline \multicolumn{1}{|c|}{Kim\cite{Kim2}} & 11.85$\pm$0.52 & 10.69$\pm$0.86 & 63.35$\pm$1.60 & 16.65$\pm$1.64 & 14.29$\pm$0.36\\
\hline \multicolumn{1}{|c|}{Tsai\cite{Tsai1}} & 21.97$\pm$1.24 & 23.91$\pm$2.46 & 40.90$\pm$1.42 & 19.45$\pm$0.73 & 17.37$\pm$0.50\\
\hline \multicolumn{1}{|c|}{Kim\cite{Kim1}} & 15.92$\pm$0.00 & 22.69$\pm$0.13 & 31.04$\pm$0.44 & 30.58$\pm$0.67 & 26.20$\pm$1.07\\
\hline \multicolumn{1}{|c|}{Ryu\cite{Ryu}} & 26.03$\pm$0.18 & 28.06$\pm$0.94 & 29.59$\pm$1.50 & 29.20$\pm$1.30 & 29.33$\pm$0.46\\
\hline \multicolumn{1}{|c|}{Proposed} & \textbf{95.81$\pm$2.51} & \textbf{89.44$\pm$3.93} & \textbf{96.09$\pm$2.37} & \textbf{92.98$\pm$2.13} & \textbf{95.25$\pm$2.71}\\
\hline
\end{tabular}
\end{center}
\label{Scaling_table}
\end{table*}

The graphical confusion matrices of the identification accuracy evaluations of all the methods are presented in Fig. \ref{Confusion}. As shown in Fig. \ref{Kim2Conf} and Table \ref{result_table}, Kim's method using halftone fingerprints had irregular identification accuracy. This was caused by similar halftone printing angles between H1 and K3, and X3 and K2. Kim's method had a limitation in that it could not identify source printers that had the same printing angles. For the proposed method, the printer identification results showed regular identification accuracy for all tested source printers. It means that the proposed method can differentiate source color laser printers that have similar halftone printing angles with high reliability.

\begin{figure}[t]
    \centerline{\includegraphics[width=8.5cm]{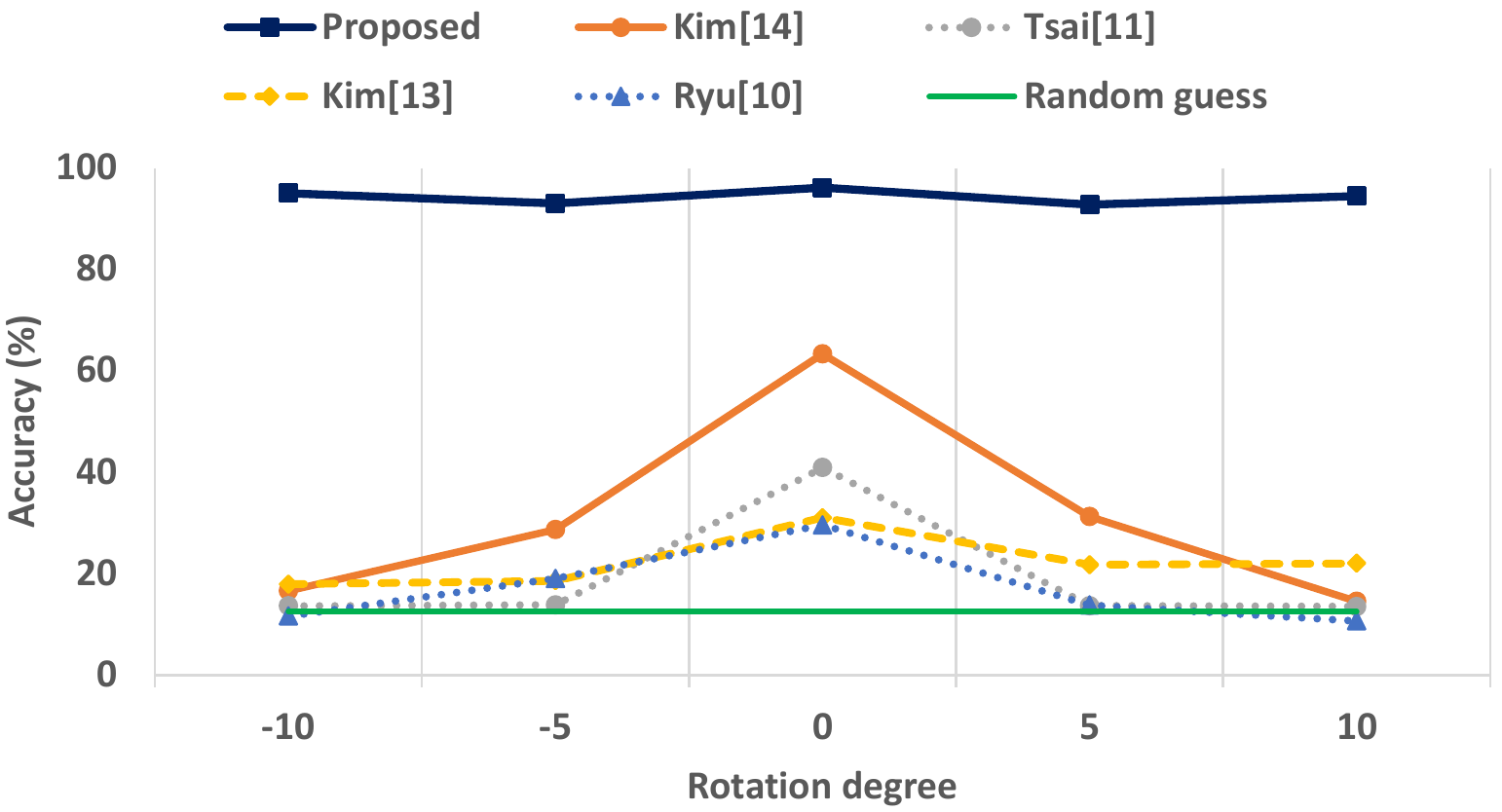}}
    \caption{Robustness evaluation results graph for rotation
    } \label{Rotation}
\end{figure}

\begin{figure}[t]
    \centerline{\includegraphics[width=8.5cm]{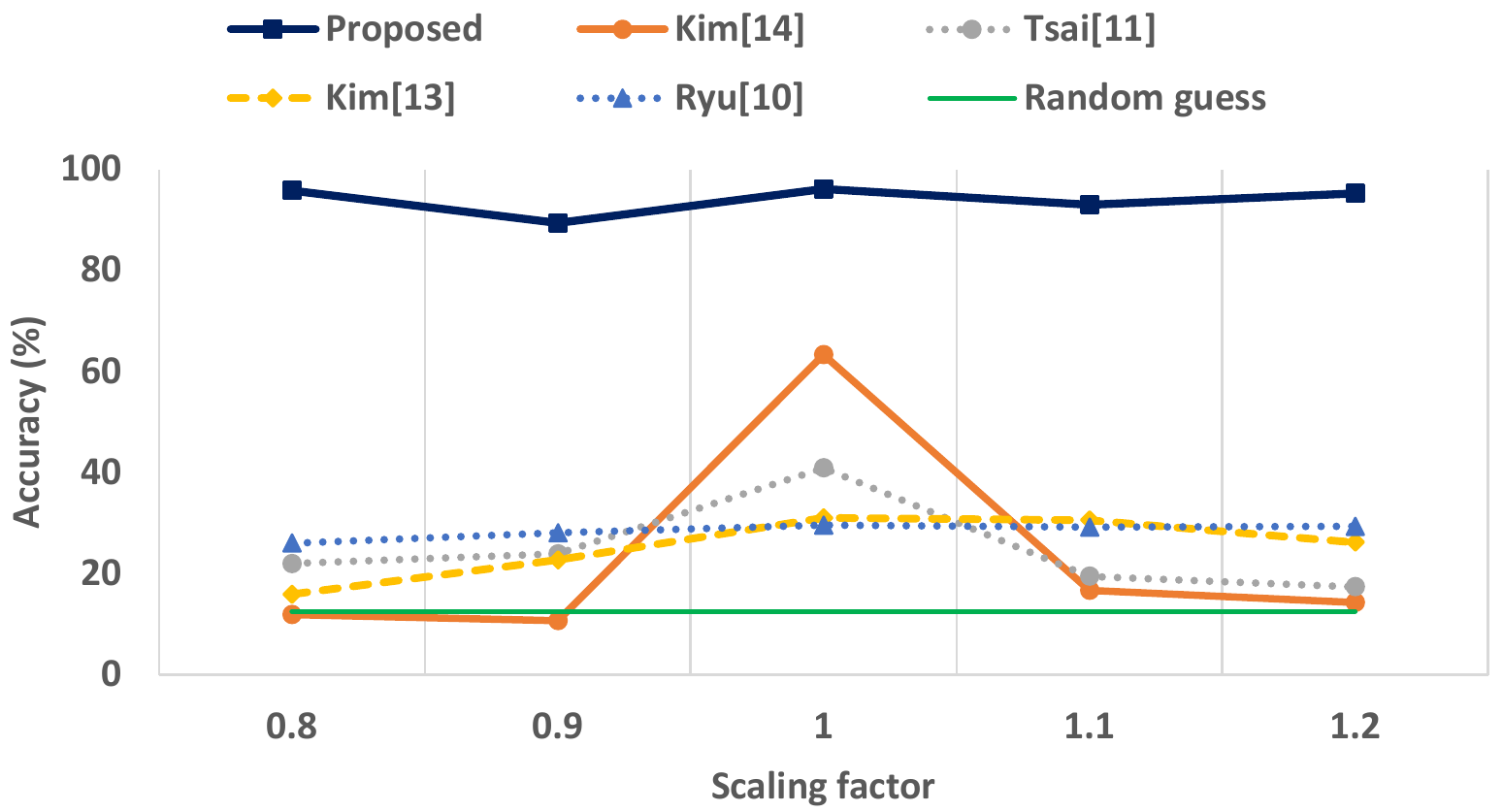}}
    \caption{Robustness evaluation results graph for scaling
    } \label{Scaling}
\end{figure}

\subsubsection{Robustness evaluation}
Robustness evaluation results are presented in Fig. \ref{Rotation}-\ref{Scaling} and Table \ref{Rotation_table}-\ref{Scaling_table}. The proposed method showed stable identification accuracy in all tests despite images that are rotated with -5 degree and +5 degree and images that are scaled with the factor of 0.9 and 1.1 were not included in the training set. It means the proposed method achieved robustness about rotation and scaling for not only trained values but also values within the trained interval.

Other methods showed performance degradation for transformed input images. Severe performance degradation occurred in all comparison methods for rotated input images. In case of scaling robustness test, Kim's method \cite{Kim1} and Ryu's method \cite{Ryu} showed stable identification accuracy because they identify the source printer mainly based on printing angle features. Scaling does not affect to the printing angle. Printing resolution feature is crucial for Kim's method \cite{Kim2} and Tsai's method \cite{Tsai1}. Thus they couldn’t identify the source printer in scaled images.

\subsection{Discussion}
We adopted a deep learning-based approach to identify the source color laser printer, and a typical limitation of deep learning is that it requires large dataset to train the network. In the experiment, only four pages of printed images were used for each source printer. Seven hundred and sixty-eight images were photographed from these printed images, and 49,152 halftone image blocks were extracted. If the source printer or several pages of printed material is available, the proposed method can be used to identify the source color laser printer. Therefore, the proposed method could be utilized in the real forensic situation despite the requirement of a large dataset.

The main limitation of the proposed method is that it cannot determine whether or not the source color laser printer of the input image is one of the candidates. In a real identification case, the source color laser printer of the input image might be none of the candidates. However, one of the trained source printers must be selected for any input image in the proposed method. To overcome this limitation, adding one more output neuron for other printers is possible. To train a network with other printers, data about other printers that includes images printed from various source printers will be necessary.

\section{Conclusion}
In this paper, we proposed a source color laser printer identification method based on cascaded learning of neural networks. Firstly, the refiner is trained to refine synthetic halftone images. Next, the HCD-CNN is trained to decompose CMYK color channels of photographed color halftone images. Based on the knowledge of the HCD-CNN, the PI-CNN is trained to identify the source printer of the input. The trained PI-CNN was used in the identification process, and the source color laser printer was selected based on the result of the PI-CNN.

Our experimental results demonstrated that the proposed method overcame the limitations of the existing methods. The proposed method achieved a state-of-the-art performance for identifying the source color laser printer of photographed input images. Since input images were taken with a smartphone with no additional close-up lens, the proposed method can be utilized to identify the source printer in a mobile environment.

For future work, we will work on reducing the computation cost. The proposed method achieved high identification accuracy and robustness about rotation and scaling, however, its computational cost is too high to operate on mobile devices. Therefore, we will test various techniques that reduce computation cost of deep neural networks and work on optimizing those techniques to our source printer identification framework.

\section*{Acknowledgement}
This work was supported by the National Research Foundation of Korea (NRF) grant funded by the Korea government (MSIT) (NRF-2016R1A2B2009595)

\end{document}